\documentclass[10pt,journal,compsoc]{IEEEtran}
\ifCLASSOPTIONcompsoc
  \usepackage[nocompress]{cite}
\else
  \usepackage{cite}
\fi
\AtBeginDocument{%
  }
    
\usepackage{balance}
\usepackage{xcolor}
\usepackage{subcaption}
\usepackage{graphicx,subcaption}
\usepackage{multirow} 
\usepackage{tikz}
\usepackage{enumitem}
\usepackage{comment}
\usepackage{pdfpages}
\usepackage{amsmath}
\usepackage{adjustbox}
\usepackage{soul}

\newcommand{\system}{Argus} % Watch2GH
\newcommand\encircle[1]{%
  \tikz[baseline=(X.base)] 
    \node (X) [draw, shape=circle, inner sep=0] {\strut #1};}
\newcommand{\parjump}{\vspace{+0.5em}}

\begin{document}

%\iffalse
%\newbool{showComments}
% \booltrue{showComments}
%\boolfalse{showComments}    % UNCOMMENT THIS BEFORE SUBMISSION
%\ifbool{showComments}{
%\newcommand{\cm}[1]{\textcolor{red}{\textbf{cm:} #1}}
%\newcommand{\jh}[1]{\textcolor{red}{\textbf{jh:} #1}}
%\newcommand{\revise}[1]{\textcolor{blue}{#1}}
\newcommand{\revise}[1]{\textcolor{black}{#1}}
%\newcommand{\deleteifnospace}[1]{\textcolor{brown}{#1}}
%}{
\newcommand{\cm}[1]{}
\newcommand{\jh}[1]{}
\newcommand{\deleteifnospace}[1]{}

%}
%\fi

\title{
%\system{}: Enabling Cross-Camera Collaboration for Video Analytics on Distributed Smart Cameras
Enabling Cross-Camera Collaboration for Video Analytics on Distributed Smart Cameras
}

\author{Chulhong~Min, Juheon~Yi, Utku Günay Acer, and
        Fahim~Kawsar%,~\IEEEmembership{Senior Member,~IEEE.}
        \IEEEcompsocitemizethanks{
        \IEEEcompsocthanksitem 
        Chulhong Min is with Nokia Bell Labs (e-mail: chulhong.min@nokia-bell-labs.com).
        \IEEEcompsocthanksitem 
        Juheon Yi is with Nokia Bell Labs (e-mail: juheon.yi@nokia.com).
        \IEEEcompsocthanksitem 
        Utku Günay Acer is with Nokia Bell Labs (e-mail: utku\_gunay.acer@nokia-bell-labs.com).
        \IEEEcompsocthanksitem 
        Fahim~Kawsar is with Nokia Bell Labs (e-mail: fahim.kawsar@nokia-bell-labs.com).
        \IEEEcompsocthanksitem 
        Fahim Kawsar is corresponding author of this paper.
        %\IEEEcompsocthanksitem 
        %This research was supported by the
}% <-this % stops an unwanted space
}

\IEEEtitleabstractindextext{%
\begin{abstract}
Overlapping cameras offer exciting opportunities to view a scene from different angles, allowing for more advanced, comprehensive and robust analysis. However, existing \revise{visual analytics} systems %\revise{\st{proposed}} 
for 
%\revise{\st{processing}} 
multi-camera streams are mostly limited to (i) per-camera processing and aggregation and (ii) workload-agnostic centralized processing architectures. In this paper, we present \system{}, a distributed video analytics system 
%\revise{\st{that runs}} 
\revise{with} \emph{cross-camera collaboration} on smart cameras. We identify multi-camera, multi-target tracking as the primary task of multi-camera video analytics and develop a novel technique that avoids redundant, processing-heavy identification tasks by leveraging object-wise spatio-temporal association in the overlapping fields of view across multiple cameras. We further develop a set of techniques to perform these operations across distributed cameras without cloud support at low latency by (i) dynamically ordering the camera and object inspection sequence and (ii) flexibly distributing the workload across smart cameras, taking into account network transmission and heterogeneous computational capacities. Evaluation of three real-world overlapping camera datasets with two Nvidia Jetson devices shows that \system{} reduces the number of object identifications and end-to-end latency by up to 7.13$\times$ and 2.19$\times$ (4.86$\times$ and 1.60$\times$ compared to the state-of-the-art), while achieving comparable tracking quality.

%It is common today to observe physical places using multiple cameras with overlapping fields of view. Recently, extensive research efforts have been made to benefit from the collaboration across such proximate cameras for wide coverage and robust monitoring. However, they still mostly treat cameras as dumb video streamers and perform the collaboration tasks in a central server, causing severe resource waste and raising massive privacy concerns. In this paper, we present \system{}, a collaborative camera system to perform robust and low-latency video analytics at edge devices. To this end, we target multi-camera, multi-target tracking as a primitive task for the cross-camera collaboration and devise a technique to intelligently filter out unnecessary, redundant identification model executions by leveraging the spatio/temporal association of target objects in overlapping cameras. We further optimise the end-to-end latency by distributing the identification operations across cameras on the fly. We evaluate \system{} with two real-world overlapping-camera datasets and two NVIDIA Jetson devices. The results show that, compared to existing solutions, \system{} reduces the number of identification model executions and end-to-end latency by up to 7.13$\times$ and 1.80$\times$ (4.86$\times$ and 1.60$\times$ compared to the state-of-the-art), while achieving the comparable tracking quality.

\end{abstract}

}

\IEEEdisplaynontitleabstractindextext

\IEEEpeerreviewmaketitle

\maketitle

\section{Introduction}~\label{sec:introduction}

\begin{table*}[pt]
\caption{\revise{Comparison of cross-camera collaboration approach in \system{} with REV~\cite{xu2022rev}, Spatula~\cite{jain2020spatula}, and CrossRoI~\cite{guo2021crossroi}.}}
%\vspace{-3pt}
\label{tab:comparison}
\begin{adjustbox}{width=1\textwidth}
\begin{tabular}{|p{0.14\linewidth}|p{0.188\linewidth}|p{0.188\linewidth}|p{0.188\linewidth}|p{0.188\linewidth}|}
\hline
& \textbf{REV}~\cite{xu2022rev} & \textbf{Spatula}~\cite{jain2020spatula} & \textbf{CrossRoI}~\cite{guo2021crossroi} & \textbf{\system{} (Ours)} \\ \hline
\textbf{Target environment} & \revise{Overlapping cameras} & Non-overlapping cameras & Overlapping cameras & Overlapping cameras \\ \hline
\textbf{Optimization goal} & \revise{On-server computation costs} & Communication and on-server computation costs  & Communication and on-server computation costs & End-to-end latency on cameras  \\ \hline
\textbf{Collaboration granularity} & \revise{Cells (group of cameras)} & Cameras  & Areas (RoIs)  & Objects \\ \hline
\textbf{Applying association} & \revise{Dynamic (depending on the target's existence)} & Dynamic (depending on the target's existence) & Static (once when the cameras are deployed) & Dynamic (depending on the target's location) \\ \hline
\textbf{Approach} & \revise{Incrementally search cells that have lowest identification confidence} & Identify the subset of cameras that capture target objects & Find the smallest RoI that contains the target objects & Minimise the \# of identification operations across cameras \\ \hline
\textbf{Video processing} & \revise{Centralized} & Centralized & Centralized & Distributed \\ \hline
\end{tabular}
\end{adjustbox}
\end{table*}
 
\begin{figure}[t]
    \centering
    \includegraphics[width=0.95\columnwidth]{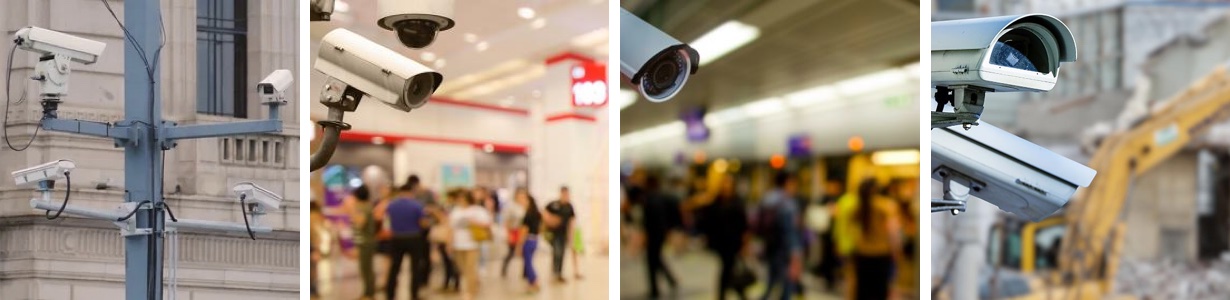}
%    \vspace{-0.1in}
    \caption{Places with overlapping cameras: intersection, shopping mall, transport, construction site.}
    \label{fig:places_cameras}
%    \vspace{-15pt}
\end{figure}

%\revise{\st{Nowadays,}} 
\revise{It is increasingly common} for physical locations to be surrounded and monitored by multiple cameras with overlapping fields of view (hereinafter 'overlapping cameras'), e.g., intersections, shopping malls, public transport, construction sites and airports, as shown in Figure~\ref{fig:places_cameras}. Such multiple overlapping cameras offer exciting opportunities to observe a scene from different angles, enabling enriched, comprehensive and robust analysis. For example, our analysis of the CityFlowV2 dataset~\cite{naphade20215aicity} (5 cameras deployed to monitor vehicles on the road intersection) shows that each individual camera separately detects \revise{only 3.7 vehicles on average}, while five cameras detect a total of 12.0 vehicles altogether. 
%\revise{Utku comment here: Is this per minute? per frame? This reads like total and it dooesn't make sense.} 
Since a target vehicle can be captured by multiple cameras from different distances and angles, we can also observe objects of interest with a holistic view. Such view diversity can make the analytics more enriched and robust, e.g., a vehicle's license plate may be occluded in one camera's view due to its position or occlusion, but not in the other cameras.

\revise{Most visual analytics systems are deployed in cloud environments. On the other hand,} \emph{on-camera video analytics} offer various attractive benefits such as immediate response, increased reliability and privacy protection. We envision that on-board AI accelerators~\cite{antonini2019resource,moss2022ultra,gong2023collaborative} (e.g., Nvidia Jetson~\cite{jetonAGX,jetonNX}, Google Coral TPU~\cite{googleCoral} and Analog MAX78000~\cite{max78000}) and embedded AI models~\cite{gordon2018morphnet, guo2021mistify} will accelerate this trend. However, the current practice of multi-camera stream processing is limited to being deployed on cameras without relying on cloud servers in two ways. (i) \emph{Per-camera processing and aggregation.} Previous work has mostly focused on processing the video analytics pipeline on each camera individually and aggregating the results at the final stage~\cite{liu2021city, shim2021multi, li2021multi}, thereby suffering from significant processing redundancy and latency. (ii) \emph{Workload-agnostic centralized processing.} Some systems have been proposed to handle enormous multi-video streams, but they mostly assume that multiple videos are streamed to the cloud and focus on optimization and coordination of the serving engine (e.g., GPU scheduling and batch processing~\cite{shen2019nexus,zhang2017videostorm}).

\revise{In this paper, we present \system{}, a distributed video analytics system designed for \emph{cross-camera collaboration} with overlapping cameras. Here, the term `cross-camera collaboration' not only encompasses the fusion of multi-view images for video analytics, but also refers to the cooperative utilization of distributed resources to ensure video analytics with high accuracy and low latency on distributed smart cameras, eliminating the need for a cloud server. To this end, we identify that \emph{multi-camera, multi-target tracking} serves as a fundamental task for multi-camera video analytics. This process involves determining the location and capturing image crops of target objects (presented as query images) on deployed cameras over time. We find that the computational bottleneck for camera collaboration arises due to the frequent execution of identification model inference across different cameras. To address this challenge, we develop a fine-grained, object-wise spatio-temporal association technique. This novel approach strategically avoids redundant identification tasks on both spatial (across multiple cameras) and temporal (within each camera over time) axes. This not only streamlines the process but also enhances the efficiency of the system.} 

To enable effective multi-camera, multi-target tracking across overlapping cameras, we develop an object-wise association-aware identification technique. Specifically, \system{} continuously tracks records of the association of objects (their bounding boxes) with the same identity across both multiple cameras (\S\ref{subsec:spatial}) and time (\S\ref{subsec:temporal}). Then, it identifies the object by matching the location association instead of running the identification model inference and matching the appearance feature. The concept of spatio-temporal association has been proposed in several previous works to reduce the repetitive appearance or query irrelevant areas~\cite{jain2019scaling,jain2020spatula,guo2021crossroi}. However, they apply to association at a \textit{coarse-grained} level, e.g., \revise{groups of cameras~\cite{xu2022rev},} cameras~\cite{jain2019scaling,jain2020spatula} or regions of interest (RoIs)~\cite{guo2021crossroi}. Thus, the expected gain is small for our target environment, which is multi-camera, multi-target tracking on overlapping cameras. For example, the resource saving from camera-wise association and filtering ~\cite{jain2019scaling, jain2020spatula} is expected to be marginal for densely deployed overlapping cameras. RoI-wise association and filtering~\cite{guo2021crossroi} also degrade tracking accuracy, as the target object is not detected on a subset of cameras. Please refer to Table~\ref{tab:comparison} and \S\ref{sec:related_work} for more details of these works. In \S\ref{subsec:exploring_opportunities} and \S\ref{subsec:overall_performance}, we also provide an in-depth analysis and a comparative study with these prior arts, respectively. Furthermore, we carefully incorporate techniques to handle corner cases in the association process (e.g., newly appearing objects, occasional failure of the identification model and its error propagation) and improve the robustness of the spatio/temporal association process (\S\ref{subsec:robustness}).

Next, we develop a set of strategies that perform spatio-temporal association over distributed smart cameras at low latency. To maximize the benefits of association-aware identification, it needs to process cameras one by one in a sequential manner so that the number of identification model inferences is minimized; identification model inference needs to be performed when the identity of the pivot object is not yet known. This would lead to an increase in end-to-end latency, even with the fewer number of identification model inferences. Also, since cameras have different workloads (i.e., the number of detected objects) and heterogeneous processing capabilities, careless scheduling and distribution might not maximize the overall performance. To this end, we develop a multi-camera dynamic inspector (\S\ref{subsec:dynamic}) that dynamically orders the camera and bounding box inspection sequence to avoid identification tasks for query-irrelevant objects. We also distribute identification tasks across multiple cameras, taking into account network transmission and heterogeneous computing capacities on the fly, to minimize end-to-end latency (\S\ref{subsec:distributing}).

We prototype \system{} on two Nvidia Jetson devices (AGX~\cite{jetonAGX} and NX~\cite{jetonNX}) and evaluate its performance with three real-world overlapping camera datasets (CityFlowV2~\cite{naphade20215aicity}, CAMPUS ~\cite{xu2016campus}, and MMPTRACK~\cite{han2023mmptrack}). The results show that \system{} reduces the number of identification model executions and the end-to-end latency by up to 7.13$\times$ and 2.19$\times$ compared to the conventional per-camera processing pipeline (4.85$\times$ and 1.60$\times$ compared to the state-of-the-art spatio-temporal association), while achieving comparable tracking quality.

We summarize the contribution of this paper as follows.

\begin{itemize}[noitemsep, topsep=4pt, leftmargin=*]
    \item We present \system{}, a novel system for robust and low-latency multi-camera video analytics with cross-camera collaboration on distributed smart cameras. 
    
    \item To enable efficient cross-camera collaboration, we develop a novel object-wise spatio-temporal association technique that exploits the overlap in FoVs of multiple cameras to optimise redundancy in the multi-camera, multi-target tracking pipeline.
    
    \item We also develop a scheduling technique that dynamically schedules the inspection sequence and workload distribution across multiple cameras to optimise end-to-end latency.
    
    \item Extensive evaluations over three overlapping camera datasets show that \system{} significantly reduces the number of identification model executions and end-to-end latency by up to 7.13$\times$ and 2.19$\times$ (4.86$\times$ and 1.60$\times$ compared to the state-of-the-art~\cite{jain2020spatula, guo2021crossroi}) while achieving comparable tracking quality to baselines.
\end{itemize} 

%The rest of the paper is organized as follows. In Section~\ref{sec:2-overview}, we first detail the overall system architecture for software-defined video analytics (Section~\ref{subsec:2-architecture}), illustrate representative app scenarios (Section~\ref{subsec:2-scenarios}), and analyze the key system requirements (Section~\ref{subsec:2-requirements}). Based on our analysis, we enumerate research issues and status quo of existing works in Section~\ref{sec:3-prior-works}. Finally, we conclude the paper in Section~\ref{sec:4-conclusion} with our future research plans.

\section{Background and Motivation}~\label{sec:background}

\subsection{Multi-Camera, Multi-Target Tracking}~\label{subsec:multi-camera-multi-target-tracking}

\revise{In this work, we focus on multi-camera, multi-tracking using deep learning-based object detection and re-identification models. These models robustly track objects across multiple views even in complex scenarios, by leveraging the discriminative power of deep neural networks. They also handle occlusions, changes in appearance, and other challenges that are difficult to address with geometry-based methods. To this end, they often learn from large-scale datasets, enabling them to generalize to a wide range of scenarios and adapt to changes in the environment.}

\begin{figure}
%\minipage{1\columnwidth}
    \centering
    \includegraphics[width=1\columnwidth]{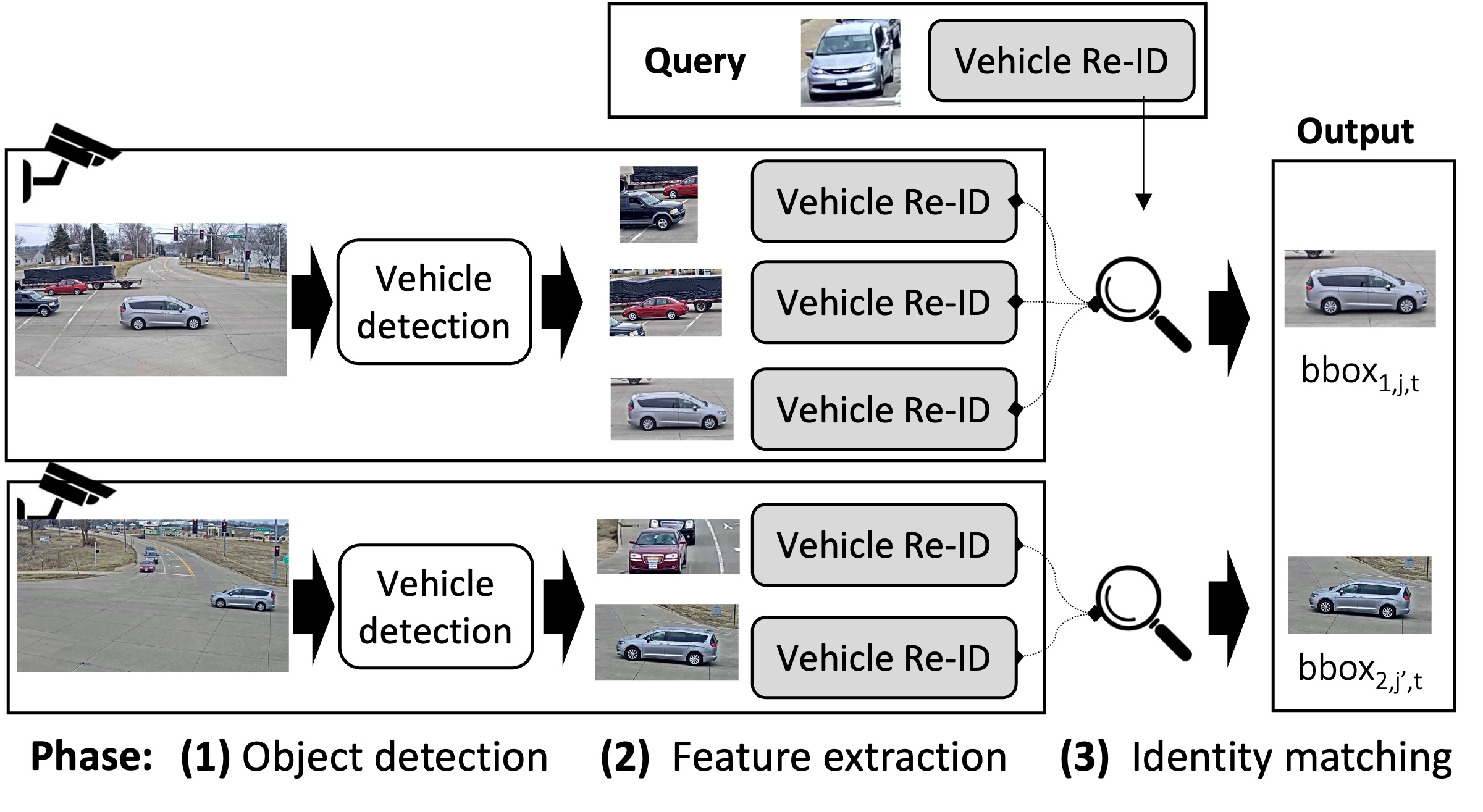}
%    \vspace{-0.15in}
	\caption{Typical pipeline for multi-camera, multi-target tracking; an example of vehicle tracking \revise{without cross-camera collaboration. Each camera runs the detection and identification independently and aggregates the output at the final stage.}}
    \label{fig:2-background-pipeline}
%\endminipage\hfill
%\vspace{-10pt}
\end{figure}

\parjump \noindent \textbf{Operational flow.}
The key to enabling video analytics on overlapping cameras is \emph{multi-camera, multi-target tracking}: detecting and tracking target objects (given as query images) from video streams captured by multiple cameras. \revise{This is typically achieved} in three stages, as shown in Figure~\ref{fig:2-background-pipeline}.
(i) The \emph{object detection} stage detects the bounding boxes of objects in one frame on each camera using object detectors (e.g., YOLO~\cite{yolov5}) or background subtraction techniques~\cite{friedman2013image,zivkovic2006efficient}.
(ii) The \emph{per-camera object identification} stage extracts the appearance features of the detected objects by running the object identification (ID) model (e.g., \cite{luo2021empirical}) and determines whether it matches the query image based on feature similarity (e.g., L2 distance, cosine similarity).
(iii) The \emph{result aggregation} stage aggregates the identification results across multiple cameras and generates tracklets~\cite{liu2021city} that can be used for further processing for application logic, e.g., object counting, license plate extraction and face recognition.

\begin{table}[pt]
%\small
\scriptsize
\centering
%\vspace{-10pt}
%\captionsetup{skip=3pt}
\caption{Identification latency on Jetson devices.}
\label{tab:motiv_benchmark}
\vspace{-5pt}
\begin{adjustbox}{width=1\columnwidth}
%\begin{tabular}{*{3}{p{\textwidth*1/7}}}
\begin{tabular}{l|c|c|c|c|c|c}
\hline
 & \multicolumn{3}{c|}{Vehicle (ResNet-101)~\cite{luo2021empirical}} & \multicolumn{3}{c}{Person (ResNet-50)~\cite{zheng2019joint}} \\ \hline
Batch size                        & 1      & 2 & 4 & 1 & 2 & 4           \\ \hline
NX               & 0.119s            & 0.206s      & 0.399s & 0.043s & 0.045s & 0.066s                         \\ \hline
AGX              & 0.065s            & 0.121s      & 0.217s & 0.018s & 0.020s & 0.028s                    \\ \hline
\end{tabular}
\end{adjustbox}
%\vspace{-5pt}
\end{table}

\parjump \noindent \textbf{Compute bottleneck: per-object identification.}
The main compute bottleneck is the execution of identification tasks, which need to be performed for all detected objects in every frame across multiple cameras to determine identity of objects, as shown in Figure~\ref{fig:2-background-pipeline}. Although we envision smart cameras equipped with built-in AI accelerators, they are not yet capable of processing a number of identification tasks in real time. Table~\ref{tab:motiv_benchmark} shows the latency of two identification models (ResNet-101-based vehicle identification~\cite{luo2021empirical} and ResNet-50-based person identification~\cite{zheng2019joint}) with different batch sizes over two Nvidia Jetson devices. It shows that the number of identification model executions to run on one camera is quite limited. For example, if 4 vehicles are detected on every frame on average, even the powerful Jetson AGX platform can only process about 4 frames per second. The throughput would drop even further if object detection is included (we show the detailed results in \S\ref{sec:evaluation}).

\subsection{Exploring Optimisation Opportunities}
\label{subsec:exploring_opportunities}

\begin{figure}[t]
    \centering
    \begin{minipage}{0.36\columnwidth}
        \centering
        \includegraphics[width=0.9\textwidth]{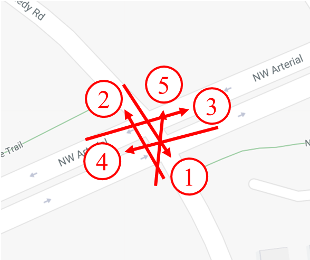}
%        \vspace{-12pt}
        \caption{Camera topology of CityFlowV2~\cite{naphade20215aicity}.}
        \label{fig:motiv_coverage_topology}
    \end{minipage}
    \hspace{1.0mm}
    \begin{minipage}{0.6\columnwidth}
        \centering
        \includegraphics[width=0.81\columnwidth]{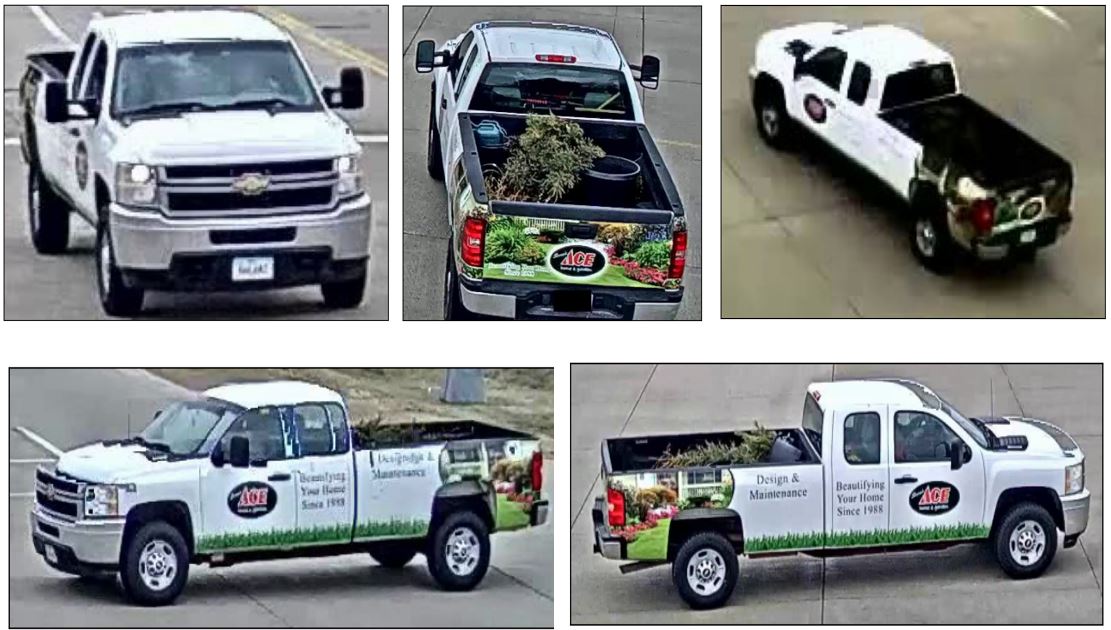}
%        \vspace{-10pt}
        \caption{The same vehicle captured from multiple views in CityFlowV2~\cite{naphade20215aicity}.}
        \label{fig:motiv_multiview}
    \end{minipage}\hfill
%\vspace{-10pt}
\end{figure}

\begin{figure}[t]
    \centering
    \begin{minipage}{0.45\columnwidth}
        \centering
        \includegraphics[width=1\columnwidth]{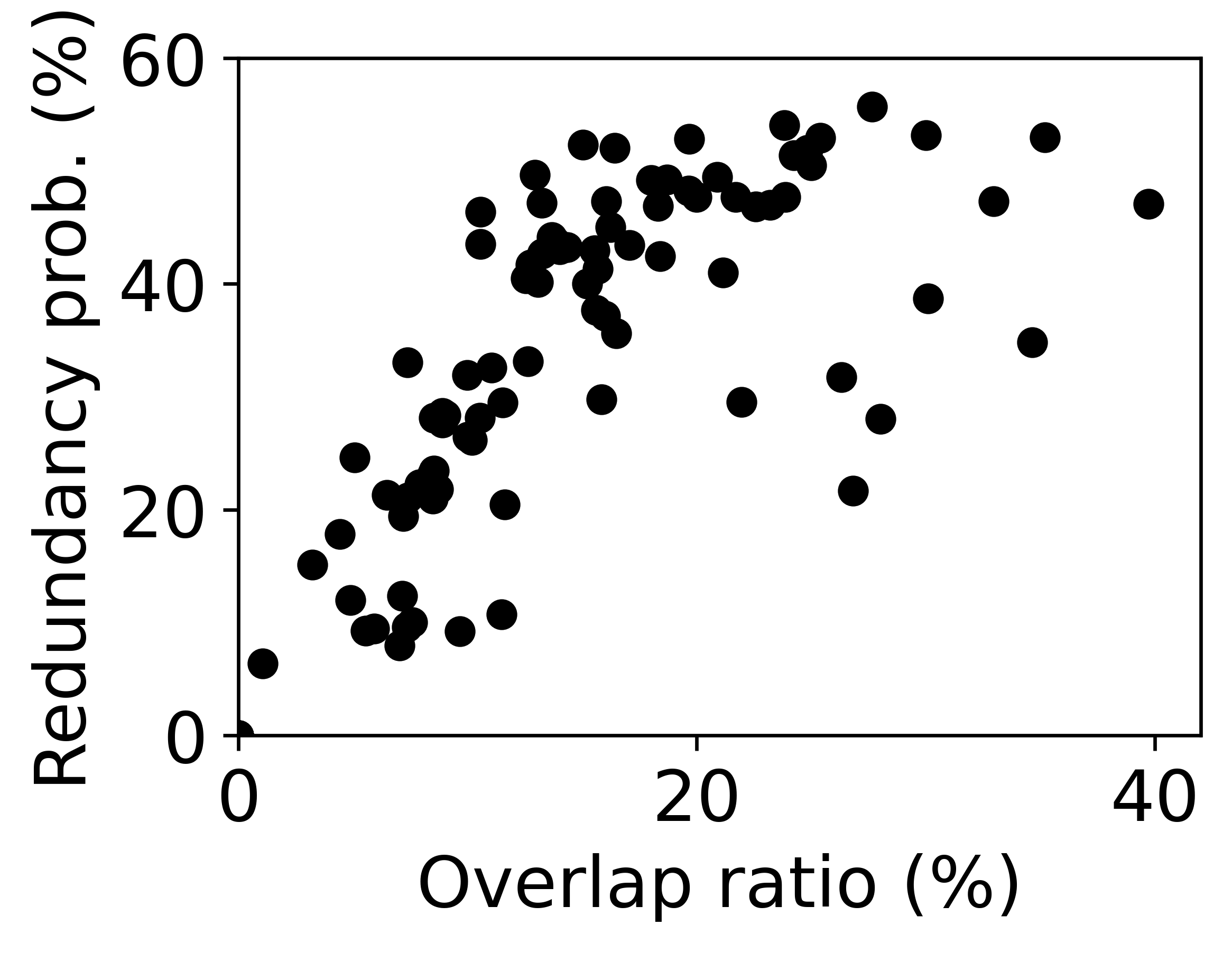}
        \vspace{-10pt}
        \caption{Identification saving opportunity for different overlapping ratios in in CityFlowV2.}
        \label{fig:motiv_reid_saving}
    \end{minipage}
    \hspace{5mm}
    \begin{minipage}{0.45\columnwidth}
        \centering
        \includegraphics[width=1\columnwidth]{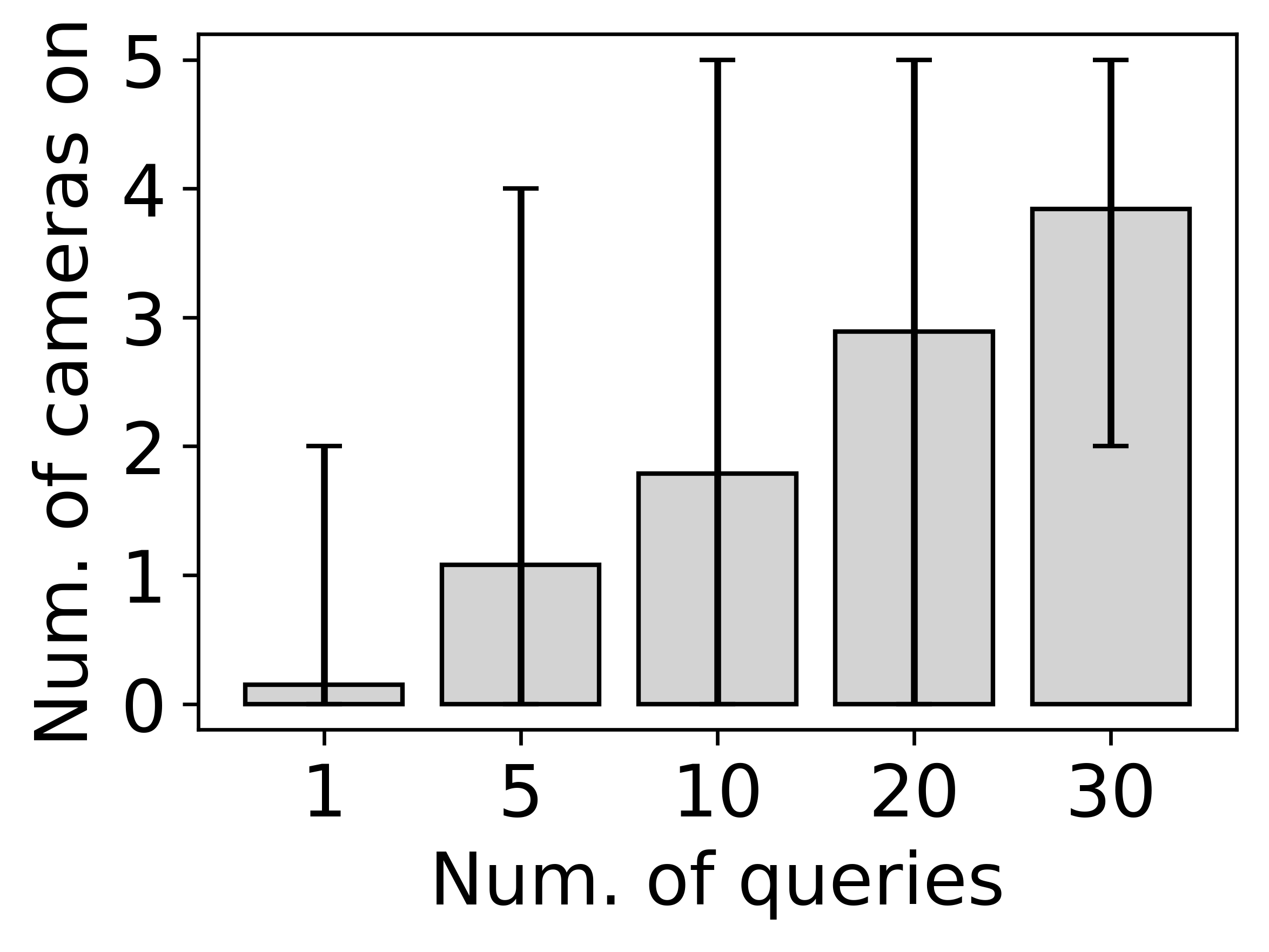}
        \vspace{-10pt}
        \caption{Number of cameras after filtering by Spatula~\cite{jain2020spatula} in CityFlowV2 (5 cameras).}
        \label{fig:motiv_spatula}
    \end{minipage}\hfill
%\vspace{-10pt}
\end{figure}

\textbf{Redundant identification of the same objects.} To explore the opportunities for optimizing the pipeline for multi-camera, multi-target tracking, we investigate the pattern of identification tasks with the CityFlowV2 dataset~\cite{naphade20215aicity}; five cameras are installed at an intersection as shown in Figure~\ref{fig:motiv_coverage_topology}. Figure~\ref{fig:motiv_reid_saving} shows the redundancy probability, i.e., the probability of objects appearing simultaneously in multiple cameras for different overlap ratios; the overlap ratio is defined as the ratio of the time the object appears simultaneously in both cameras and the total time it is detected in any camera; for a target appearing in $n$ cameras, we calculate all pairwise overlap ratios ($nC_2$) and take the average. %\revise{Utku comment: I don't understand this last sentence. The redundancy probability and overlap ratio is not explained properly.} 
Each point represents a different query. The results show that, as the overlap ratio increases, the probability of an object's appearance in multiple cameras also becomes higher. This means that a dense array of cameras with overlapping FoVs will have more redundant identification tasks for the same object across multiple cameras.
%\revise{Utku comment: This whole paragraph needs to re-written.}

\begin{figure}[t]
    \centering
    \begin{minipage}{0.55\columnwidth}
        \centering
        \includegraphics[width=1.0\columnwidth]{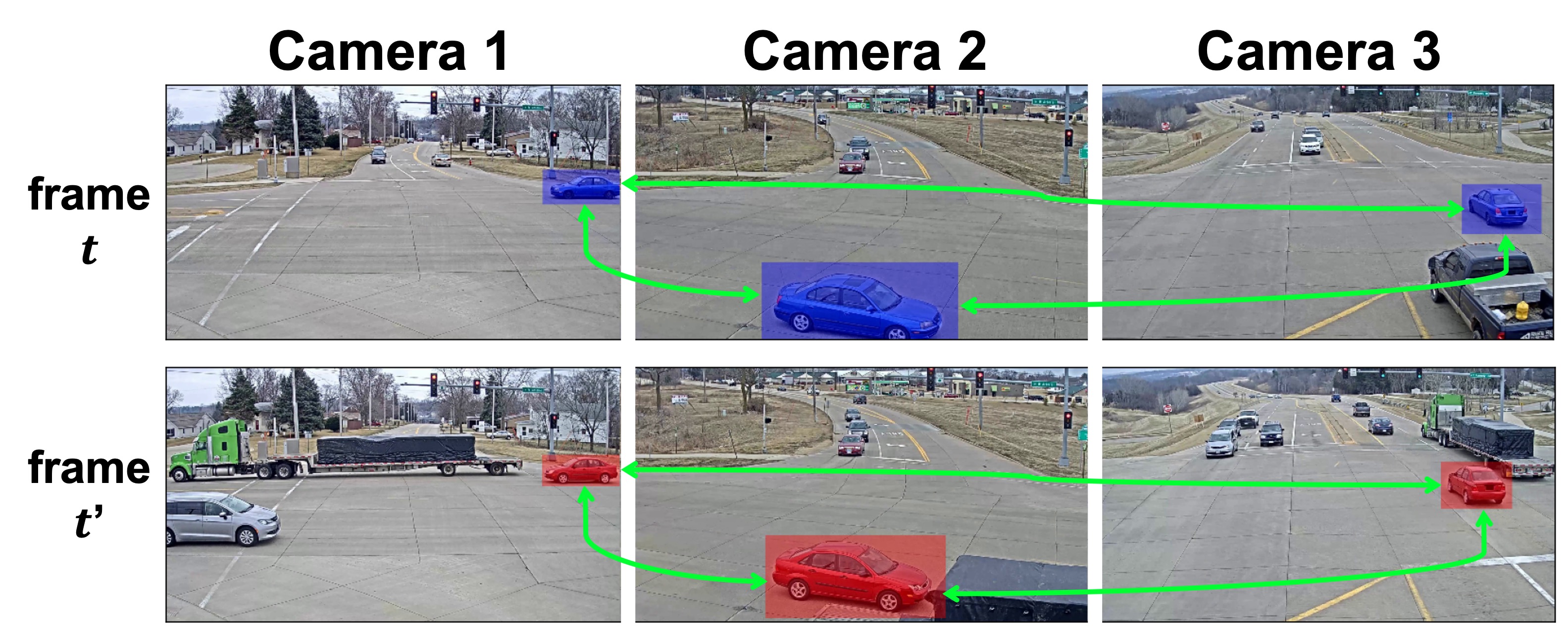}
%        \vspace{-10pt}
        \caption{Example of spatial association (green lines).}
        \label{fig:system_spatial_association}
    \end{minipage}
%    \hspace{5mm}
    \begin{minipage}{0.43\columnwidth}
        \centering
        \includegraphics[width=0.95\columnwidth]{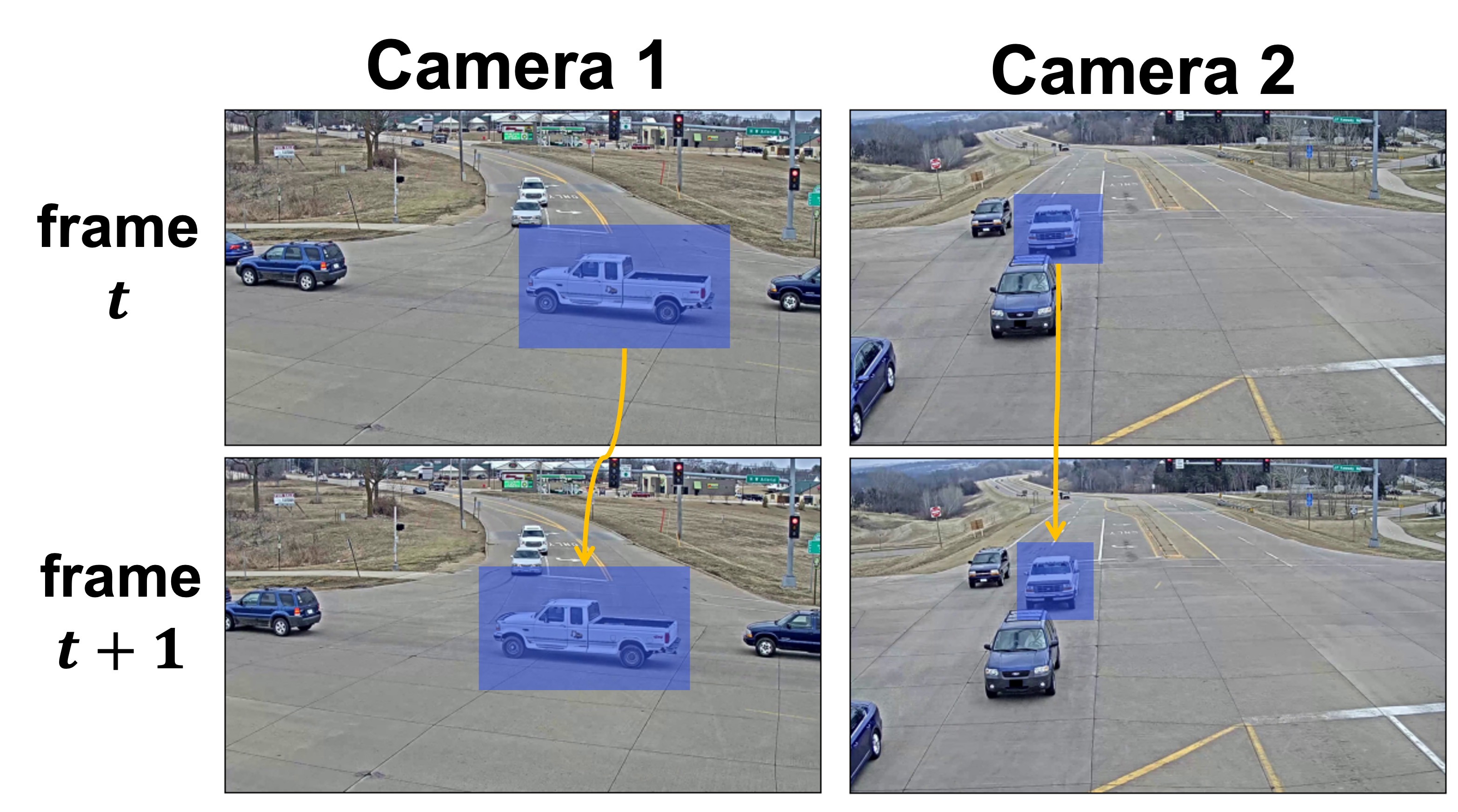}
%        \vspace{-10pt}
        \caption{Example of temporal association (yellow lines).}
        \label{fig:system_temporal_association}
    \end{minipage}\hfill
%\vspace{-10pt}
\end{figure}

\parjump \noindent \textbf{Spatio-temporal association.} \revise{To avoid unnecessary and redundant identification tasks, we adopt \emph{spatio-temporal association} of objects, which have been proposed in the auto-calibration techniques~\cite{hartley2003multiple,stein1999tracking} for multi-view tracking systems.} Spatial-temporal association refers to the geographical and temporal association of an object to different cameras. More specifically, we associate the identity of an object across multiple cameras by matching their correlated positions on the frame, rather than matching appearance features extracted from the identification model, as shown in Figure~\ref{fig:2-background-pipeline}. This intuition arises from the observation that, once installed in a place, cameras' FoVs are fixed over time. We explain spatial association with an example. If the bounding box of two objects (at different times) is located at the same position in one camera's FoV, the position of their bounding boxes in other cameras will also remain the same.\footnote{Of course, this argument is not always right in theory. Since a camera projects 3D space onto the 2D plane, the same bounding box of one camera at different times does not guarantee the same position of an object. The simplest case would be when two objects of different sizes are located in the same direction from the camera but a smaller object is located closer to the camera. However, in practice, such cases are very rare because the camera is often installed to look at a 2D plane (e.g., street and floor) obliquely to cover a wide area and objects of interest (e.g., vehicles and people) cannot be located at arbitrary 3D positions, as shown in Figure~\ref{fig:system_spatial_association}. Also, although such a case happens (e.g., two objects at different positions are captured in the same bounding box in Camera 1), the spatial association of two objects is not made because the position and size of the bounding boxes in other cameras (e.g., Camera 2 and 3) will be different.} Figure~\ref{fig:system_spatial_association} shows the spatial association obtained from the CityFlowV2 dataset~\cite{naphade20215aicity}. Each row shows a list of images captured by three cameras (Camera 1, 2, 3) installed as in Figure~\ref{fig:motiv_coverage_topology} at the same time. Each column shows the images taken by the same camera. The red and blue overlay boxes in each row represent the bounding box of a red vehicle and a blue vehicle, respectively. Although two vehicles crossed the intersection at different times, when two vehicles are located at a similar location on Camera 1, we can observe that the corresponding bounding boxes remain in a similar position on the other cameras. Similarly, as shown in Figure~\ref{fig:system_temporal_association}, we expect the \emph{temporal association} of an object, which means that an object in a video stream remains in proximity within successive frames.

\subsection{Limitations of Prior Work} 
\label{subsec:prior_work_limitation}

Despite such benefits, developing an effective filtering strategy using spatio/temporal association is not straightforward. There are prior works that explored spatio/temporal association for filtering out redundant workloads from multiple video streams. We present their techniques and limitations; we explain their work in more detail in \S\ref{sec:related_work}.

\parjump \noindent \textbf{\revise{Auto calibration using spatio/temporal association.}} \revise{Auto-calibration, also known as self-calibration, has been proposed as a solution to enable multi-view tracking systems from 1990s. This technique aims to automatically estimate the camera parameters, such as intrinsic and extrinsic parameters for object tracking in multiple camera views, without the need for manual intervention or specialized calibration objects~\cite{hartley2003multiple}. Auto-calibration methods leverage the spatio-temporal correlation of objects in multiple views as described in \S\ref{subsec:exploring_opportunities}, taking advantage of the geometric constraints imposed by the scene and the motion of objects or the camera itself~\cite{pollefeys2004visual}. By harnessing these constraints, auto-calibration techniques can iteratively refine the camera parameters, leading to improved tracking accuracy and robustness~\cite{sturm1996factorization}. Several auto-calibration methods have been proposed in the literature, including the self-calibration of space and time technique~\cite{stein1999tracking}, which exploits the correlation between space and time in the image sequence to estimate the camera parameters. Other approaches~\cite{makris2004bridging,black2002multi} utilize the epipolar geometry and geometric constraints to estimate the camera parameters. Additionally, the establishment of a common coordinate frame across multiple views has been proposed to improve tracking performance~\cite{lee2000monitoring}. While auto-calibration methods have shown the feasibility of object tracking from multiple camera views, they still face several limitations, especially when compared to modern approaches that utilize deep learning-based object detection and identification models. Auto-calibration methods are typically based on geometric constraints, which can be sensitive to errors in feature detection and correspondence matching, leading to inaccurate camera parameter estimation. Also, these methods rely on the assumption of a static scene, which may not hold true in dynamic environments where objects and people are constantly moving and changing~\cite{stein1999tracking}. Moreover, these techniques struggle with real-deployment environments due to computational constraints, as the complexity grows with the number of cameras and tracked objects. In this paper, we propose a novel approach that enables multi-camera, multi-object tracking on cameras using deep learning-based object detection and identification models.}

\parjump \noindent \textbf{\revise{Camera-wise filtering in non-overlapping cameras.}} Spatula~\cite{jain2019scaling, jain2020spatula} leverages cross-camera correlation to identify a subset of cameras likely to contain the target objects and filter out unnecessary cameras (that do not contain the target objects). While it shows a significant performance benefit in its target environment (widely deployed \emph{non-overlapping} cameras), it fails to effectively reduce redundant identification operations in \emph{overlapping} cameras. To quantify its benefit, we analyzed the CityFlowV2 dataset~\cite{naphade20215aicity}. Figure~\ref{fig:motiv_spatula} shows the average number of cameras out of five cameras, used by Spatula; the error bar indicates the minimum/maximum number of cameras. The results show that the benefit of Spatula-based camera-wise filtering quickly diminishes when more queries are used, i.e., fewer cameras are filtered out. This is because a higher number of objects are likely to be captured by a higher number of cameras, simultaneously.

\parjump \noindent \textbf{\revise{Camera-wise filtering in overlapping cameras.}}
\revise{REV~\cite{xu2022rev} leverages spatial correlation across multiple overlapping cameras to minimize the number of processed cameras in identifying the target object. However, its goal is to \textit{confirm the presence} of the target object within a given timestamp. As such, it cannot be applied for \system{}, which not only aims to \textit{confirm the presence} of a target object but also \textit{extract the image crops of the target} from all cameras that capture it. Specifically, REV employs an incremental approach, starting its search from the camera that detects the significant number of objects~\footnote{The underlying rationale is that cameras with more bounding boxes are more likely to capture the target object.} and discontinues the search once the target is identified. Thus, it often misses the image crops from the remaining cameras, which may have captured the target in superior quality.}

\begin{figure}[t]
    \centering
    \begin{subfigure}[b]{0.475\columnwidth}
        \centering
        \includegraphics[width=1\textwidth]{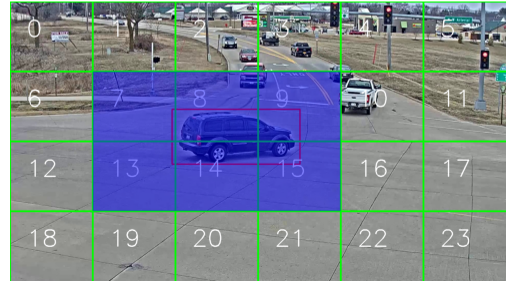}
%    \vspace{-0.1in}
    \caption{Camera \#1 (RoI: 6 grids).}
    \label{fig:motiv_crossRoI_1}
    \end{subfigure}
    %\hspace{0.002\textwidth}
%    %\hspace{1mm}
    \begin{subfigure}[b]{0.475\columnwidth}
        \centering
        \includegraphics[width=1\textwidth]{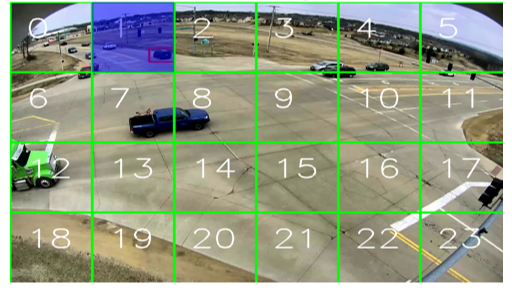}
%    \vspace{-0.1in}
    \caption{Camera \#2 (RoI: 1 grid).}
    \label{fig:motiv_crossRoI_2}
    \end{subfigure}
    \vspace{-0.1in}
    \caption{Overlapping RoI example between 2 cameras. CrossRoI~\cite{guo2021crossroi} favours Camera \#2 which has the smaller RoI size.}
    \label{fig:motiv_crossRoI}
%\vspace{-15pt}
\end{figure}

\parjump \noindent \textbf{\revise{RoI-wise filtering in overlapping cameras.}} CrossRoI~\cite{guo2021crossroi} leverages spatio-temporal correlation to optimize the region of interest (RoI) of multiple video streams from overlapping cameras. When multiple objects are captured by a set of cameras from different views, CrossRoI extracts the smallest possible total RoI across all cameras in which all target objects appear at least once, and then reduces processing and transmission costs by filtering out unmasked RoI areas, i.e., (a) redundant appearances and (b) areas that do not contain the target objects; RoI is defined as a 6-by-4 grid. While it effectively reduces the workload to be processed, it is not suitable for multi-camera, multi-task tracking. Since it aims to minimize the RoI size that covers the overlapping FoVs, the smaller RoI that contains the object is preferred (e.g., 1 grid in Camera 2 instead of 6 grids in Camera 1 in Figure~\ref{fig:motiv_crossRoI}). This would lead to considerable degradation of the accuracy. Furthermore, since it filters out redundant appearance in the initial stage, analytics applications cannot benefit from a holistic view, as shown in Figure~\ref{fig:motiv_multiview}.
\section{\system{} Design}
\label{sec:design}

\subsection{Design Goals}
\noindent \textbf{Low-latency and high accuracy.} We aim at achieving both low latency and high accuracy in running multi-camera, multi-target tracking across overlapping cameras, which is the key requirement of various video analytics apps. 

\parjump \noindent \textbf{On-device processing on distributed smart cameras:} 
Streaming videos to a cloud server for processing incurs significant networking and computing costs as well as privacy issues. We aim to run the video analytics pipeline with cross-camera collaboration fully on distributed smart cameras leveraging on-device resources.

\parjump \noindent \textbf{Flexibility of tracking pipeline.} 
We treat the AI models as a black box, thereby supporting both open-source and proprietary models and allowing analytics app developers to select the models for the purpose flexibly. 

\subsection{Approach}

\textbf{Multi-camera object-wise spatio-temporal association.} Our preliminary study reveals that the computational bottleneck for multi-camera, multi-target tracking in overlapping cameras is the redundant identification of the same object (\S\ref{subsec:multi-camera-multi-target-tracking}). To achieve both resource-efficient and accurate tracking, we devise a method for \emph{object-wise association-aware identification}. As shown in Figures~\ref{fig:system_spatial_association} and \ref{fig:system_temporal_association}, \system{} associates the spatio-temporal correlation of \emph{objects}' positions and identifies redundant \emph{identification} tasks. It reduces on-camera computational costs by filtering out redundant identification tasks for the same object across multiple cameras (spatially) and over time (temporally). It also provides accurate tracking by guaranteeing tracking information on all cameras.

\parjump \noindent \textbf{On-camera distributed processing.} To enable on-camera processing, we further devise two optimization techniques. First, we optimize on-camera workload by minimizing the number of model executions (both object detection and identification). By inspecting cameras and objects (bounding boxes) in order of probability of containing the target object, \system{} avoids model executions that are irrelevant to the target objects; note that the tracking operation is finished when all target objects are found. Second, we further optimize end-to-end latency with parallel execution on distributed cameras. More specifically, \system{} distributes the identification workload across multiple cameras on the fly and executes it in parallel.

\begin{figure}%[t]
\minipage{1\columnwidth}
    \centering
    \includegraphics[width=1\textwidth]{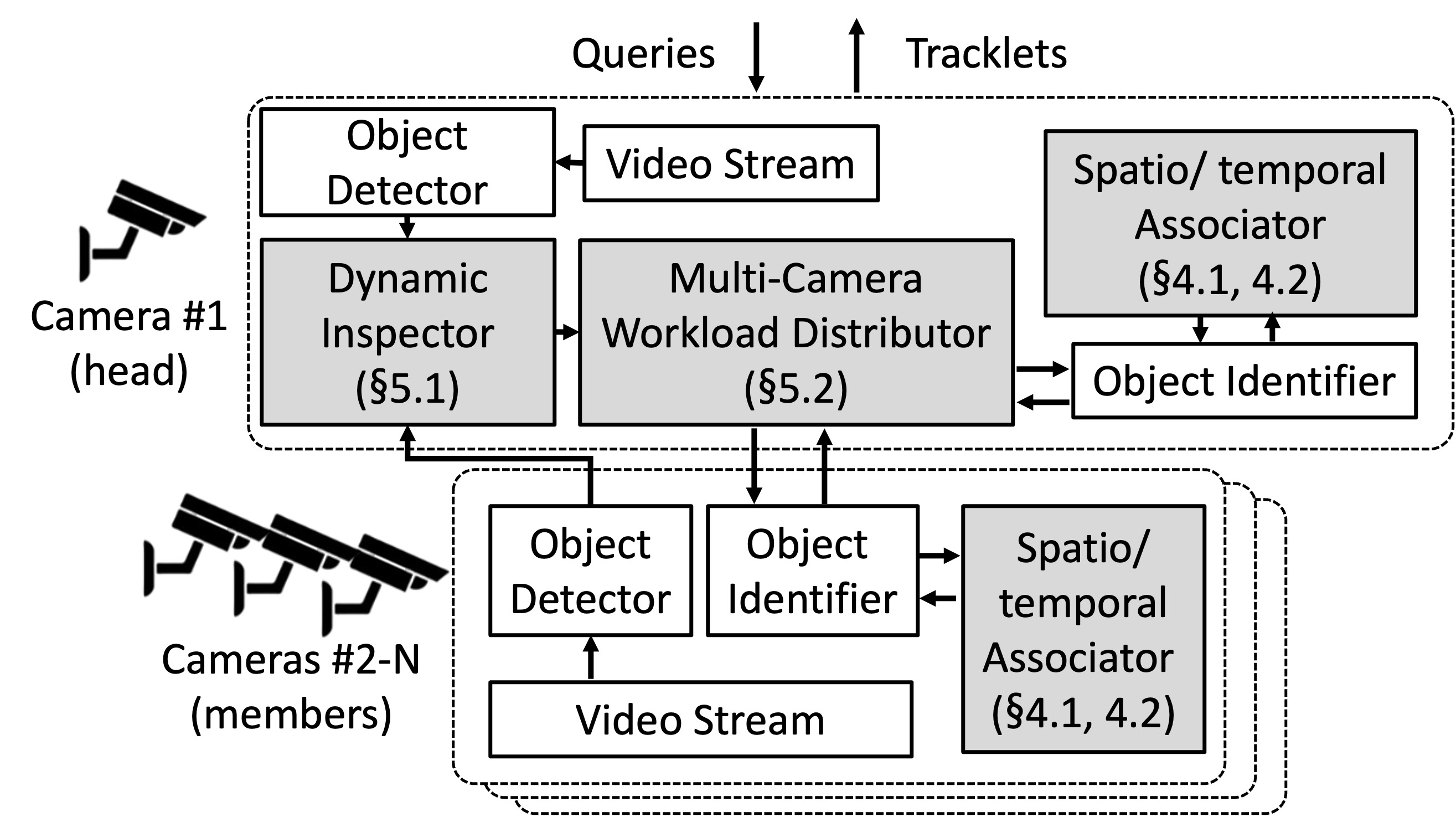}
	\caption{\system{} system architecture.}
    \label{fig:system_architecture}
\endminipage\hfill
\vspace{-5pt}
\end{figure}

\subsection{System Architecture}

Figure~\ref{fig:system_architecture} shows the system architecture of \system{}. It takes the query images as input from analytics apps and provides the tracklets (list of cropped images and bounding boxes of the detected objects) tracked from multiple cameras as output. Given the targets to track, \system{} first starts by running the object detector to detect objects for identification on each frame in parallel. Afterwards, the head camera runs the Dynamic Inspector (Section~\ref{subsec:dynamic}) to determine the processing order of cameras and bounding boxes. Once the processing order is determined, the Multi-Camera Workload Distributor (Section~\ref{subsec:distributing}) schedules the identification tasks across cameras (head and members), considering the network transmission latency and heterogeneous compute capabilities. Given the identification workloads, each camera runs the object identifier; the Spatio-temporal Associator (Sections~\ref{subsec:spatial} and \ref{subsec:temporal}) opportunistically skips the inference by leveraging the spatio-temporal correlations across cameras.

%\section{Multi-Camera Spatio/Temporal Association}
\section{Spatio-Temporal Association}
\label{sec:spatialtemporal}

\begin{figure}%[t]
\minipage{1\columnwidth}
    \centering
    \includegraphics[width=1\textwidth]{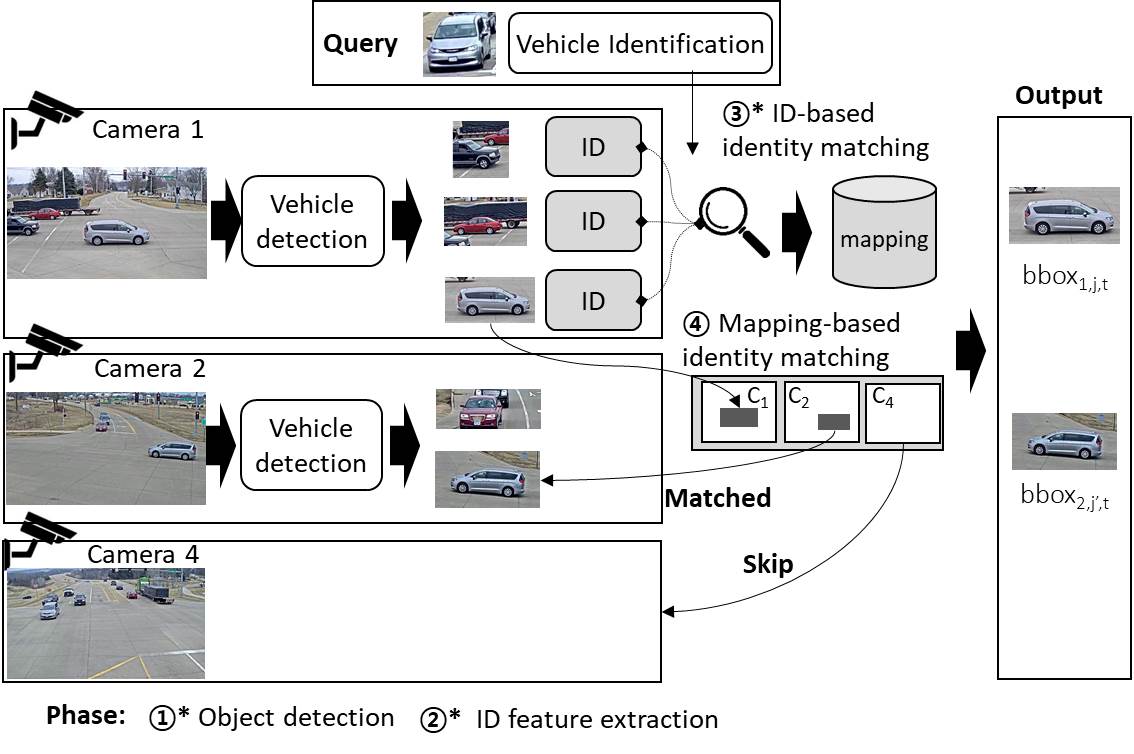}
	\caption{\revise{Overview of our cross-camera collaboration-enabled multi-camera spatio-temporal association. (*): operations are performed on a subset of cameras.}}
    \label{fig:system_approach}
\endminipage\hfill
%\vspace{-10pt}
\end{figure}

\parjump \noindent \textbf{Operation Overview.} 
\revise{The goal of multi-camera spatio-temporal association is to accurately track the query identities from multiple cameras with a minimal number of identification operations, which is the key bottleneck of the multi-camera, multi-object tracking pipeline.} Figure~\ref{fig:system_approach} shows the operation of multi-camera spatio-temporal association, illustrating how to leverage the association to achieve efficient multi-camera, multi-target tracking. For example, if an object matching the query is found in Camera 1 and the \textit{expected} position of its bounding box in Camera 2 can be obtained, we determine the identity of an object in Camera 2 if its position matches the expected position. If no bounding box that corresponds to the one in Camera 1 is expected to exist in other cameras, e.g., in Camera 4, we skip all identification operations in Camera 4, as this means that the object is located outside of Camera 4's FoV.

\parjump \noindent \textbf{Formulation.}
We formalize our problem setting as follows:
\begin{itemize}[leftmargin=*]
	\setlength\itemsep{-0.1em}
	\item $C$: a set of cameras, where $C^i$ is $i^{th}$ camera,
	\item $F_t$: a set of image frames at time $t$, where $F_t^i$ is an image frame from $C^i$ at time $t$.  
	\item $E_Q$: a set of id feature embedding of query images, where $E_j$ is the feature embedding of $j$th query 
        \item \revise{{$bbox_{t,j}^i$}}: bounding boxes of the detected on $C^i$ at time $t$. $C^i$ has $n_t^i$ objects detected at time $t$ ($j=1,2,...,n_t^i$).
\end{itemize}

\revise{Formally, the goal of multi-camera spatio-temporal association is to minimize the total number of identification operations across all cameras,}

\begin{equation}
\min \sum_i n_{IDs}^i,
\end{equation}

\noindent \revise{where $n_{IDs}^i$ is number of identification operations on $C^i$.}

\parjump \noindent \textbf{Operational flow.} \system{} operates as follows in detail. For simplicity, we explain the procedure for a single query.

\begin{enumerate}[leftmargin=*]
	\setlength\itemsep{-0.1em}
	\item \textbf{Camera order determination.} We decide the order of cameras to inspect (\S\ref{subsec:dynamic}), and repeat below steps for each camera.
	\item \textbf{Object detection (\encircle{1} in Figure~\ref{fig:system_approach})}: For a frame from $i$-th camera $C^i$, we perform the object detection. We define its output as $\{bbox_{t,j}^i, label_{t,j}^i\}$, where $bbox_{t,j}^i$ and $label_{t,j}^i$ are a bounding box and a label for $j$-th object on $C^i$ at time $t$, respectively.
	\item \textbf{ID feature extraction (\encircle{2} in Figure~\ref{fig:system_approach}).} For objects with a detected label that matches the query label, we sort the bounding boxes for inspection (\S\ref{subsec:dynamic}). For each cropped image, we execute the identification model and obtain the ID features, $\{E_{t,j}^i\}$, where $E_{t,j}^i$ is the ID appearance feature from the cropped image ($bbox_{t,j}^i$). We further reduce the identification operations within a camera by leveraging the temporal locality of an object (\S\ref{subsec:temporal}).
	\item \textbf{Identity matching (\encircle{3} in Figure~\ref{fig:system_approach}).} For each object, we compute its similarity to a query image by comparing their extracted features, \revise{$E_Q$}, and determine its identity. Steps 3--4 are repeated until the target object is found.
	\item \textbf{Mapping-based identity matching (\encircle{4} in Figure~\ref{fig:system_approach}).} We construct a set of bounding boxes of the target object from previously inspected cameras including the current camera, $C_i$, i.e., $\{entry\_bbox_{t,j}^k\ \mid k \subset K\}$, where $K$ is a set of cameras that are inspected. Note that it may contain one or more \textit{N/A} elements, which show that the object is located in non-(or partially) FoV of $C_i$ at time $t$. We look up the mapping entry that matches $\{entry\_bbox_j^k\}$ for the camera set, $K$. If the entry is found, then we extract the bounding boxes of other cameras (that are not inspected yet) in the entry, i.e., $\{entry\_bbox_j^i \mid i \notin K \}$.
	\begin{enumerate}
		\item If the bounding box in other camera exists in the entry, e.g., $entry\_bbox_{t,j}^{i'}$, we perform object detection in the corresponding camera, $C^{i'}$ and determine the query identity by spotting the bounding box that matches $entry\_bbox_{t,j}^{i'}$.
		\item If the entry has \textit{N/A} in other cameras, e.g., $C^{i''}$, we skip all the operations of $C^{i''}$ at time $t$.  
	\end{enumerate}
	\item If a target object is not found in the frame, we set $entry\_bbox_t^j$ as \textit{N/A} and do step 5. 
	\item Steps 1--6 are repeated until all cameras are inspected.
\end{enumerate}

When multiple queries are given, the output from object detection \encircle{1} and ID feature extractions \encircle{2} is shared, but identification and mapping-based identity matching (\encircle{3} and \encircle{4}) are performed separately. We explain the implementation details in \S\ref{subsec:implementation}.

\subsection{Spatial Association}
\label{subsec:spatial}

We first explain how we define a spatial association across multiple cameras. Once an object with the same identity is captured by multiple cameras, we create a mapping entry that contains a timestamp and a list of the corresponding bounding boxes on each camera in $C$. We use bounding boxes as location identifiers for fine-grained matching of the spatial association. Formally, we define a mapping entry as $entry^j=\{entry\_bbox_j^i\}$, where $entry\_bbox_j^i$ is a pair of coordinates referring to the southwestern and northeastern corner of the box in $C^i$ at the $j$th mapping entry. $entry\_bbox_j^i$ is set to \textit{N/A} if the object is not found in the corresponding camera, $C^i$.

\noindent \revise{
\textbf{Utilizing the spatial association.} In subsequent time intervals, we apply the identification model to the detected objects in a single camera (please refer to \S\ref{subsec:dynamic} for determining the order of camera inspection). Upon identifying an object, we search for a mapping entry matching a bounding box of the identified object in the same camera. If such an entry is found, we examine the detected bounding boxes on the remaining cameras whose entry value is not \textit{N/A}. If a bounding box in the remaining camera matches the located entry, we associate (i.e., reuse) the identification result from the first camera, avoiding the need to rerun the identification model. Note that searching for a matching entry in the mapping table involves calculating the bounding box overlap, which is an extremely lightweight operation (e.g., takes <1~ms for 1,000 matches) as detailed in \S\ref{subsec:implementation}.
}

\noindent \revise{\textbf{Management of the spatial association.} We use bounding boxes as location identifiers for fine-grained matching of the spatial association. To facilitate quick access, we maintain the entries as a hash table. Also, if the number of entries exceeds a threshold (e.g., 100), \system{} filters out duplicate or closely located entries by running non-maximum suppression on the bounding boxes of the entries. Specifically, when two entries have bounding boxes from the same cameras with significant overlap, we retain only the entry that has (i) a higher number of non-\textit{N/A} values and (ii) a higher average identity matching score; implementation details are provided in \S\ref{subsec:implementation}. These mapping entries can be obtained during the offline phase with pre-recorded video clips or updated during the online phase with runtime results.}

\subsection{Temporal Association}
\label{subsec:temporal}

We leverage temporal association to further reduce the number of identification operations. It is inspired by the observation proposed in simple online tracking methods~\cite{bewley2016simple,wojke2017simple}, that the location of an object does not change significantly within a short period of time. That is, the bounding box of an object in a video stream would remain in proximity to the bounding box with the same identity in the previous frame. For example, even in the vehicle tracking scenario in CityFlowV2~\cite{naphade20215aicity}, the distance of a vehicle moving at a speed of 60 km/h in successive frames of a video stream at 10 Hz is about 1.7 metres, which is relatively small compared to the size of the area that a security camera usually covers.

When ID feature extraction is performed, \system{} caches the ID features with their bounding box. Then, when the ID feature is needed for a new bounding box in a later frame, \system{} finds the matching bounding box in the cache; we explain the implementation details of bounding box matching in \S\ref{subsec:implementation}. When the matching bounding box is found, \system{} reuses its ID feature and updates the bounding box in the cache. We set the expiry time to one frame, i.e., the cache expires in the next frame unless it is updated.

\subsection{Implementation of Association Technique~\label{subsec:implementation}}

\noindent \textbf{RoI extraction.} There are several options for the RoI extraction stage that can be adopted in \system{} (e.g., background subtraction~\cite{friedman2013image,zivkovic2006efficient} or object detection models~\cite{yolov5}). Although the background subtraction method is more lightweight, we use the object detection method because the object detection method can effectively reduce the number of ROIs to be examined by matching the corresponding labels with the object class of the query (e.g., vehicles or people). In this paper, we use the YOLOv5n model, the lightest model in the YOLOv5 family~\cite{yolov5} as it provides reasonable detection accuracy even for small cropped images in 1080p streams of our datasets. Note that app developers can flexibly use different RoI detection methods depending on the processing capacity of smart cameras and the service requirements.

\parjump \noindent \textbf{Identity matching.}
For identification, it is common to train the object type-specific identification models (e.g., vehicle and person) and establish correspondences by measuring the similarity between the feature vectors of the (cropped) images (e.g., Euclidean distance or cosine similarity). We use the dataset-specific identification models and similarity functions to ensure the accuracy of tracking (details in \S\ref{subsec:eval_setup}).

\parjump \noindent \textbf{Bounding box matching.}
The key to leveraging the spatio/temporal association is to match the bounding box of a detected object with the other bounding boxes in the mapping entry and in the previous frame. We use the intersection-over-union (IoU) to measure the overlap between two bounding boxes and detect a match if the IoU value exceeds 0.5 (widely used threshold for object tracking~\cite{everingham2010voc}). \revise{Note that IoU calculation overhead is negligible (e.g., takes $<$1~ms even for 1,000 matchings on Jetson AGX board).}

\begin{figure}
    \centering
    \includegraphics[width=1\linewidth]{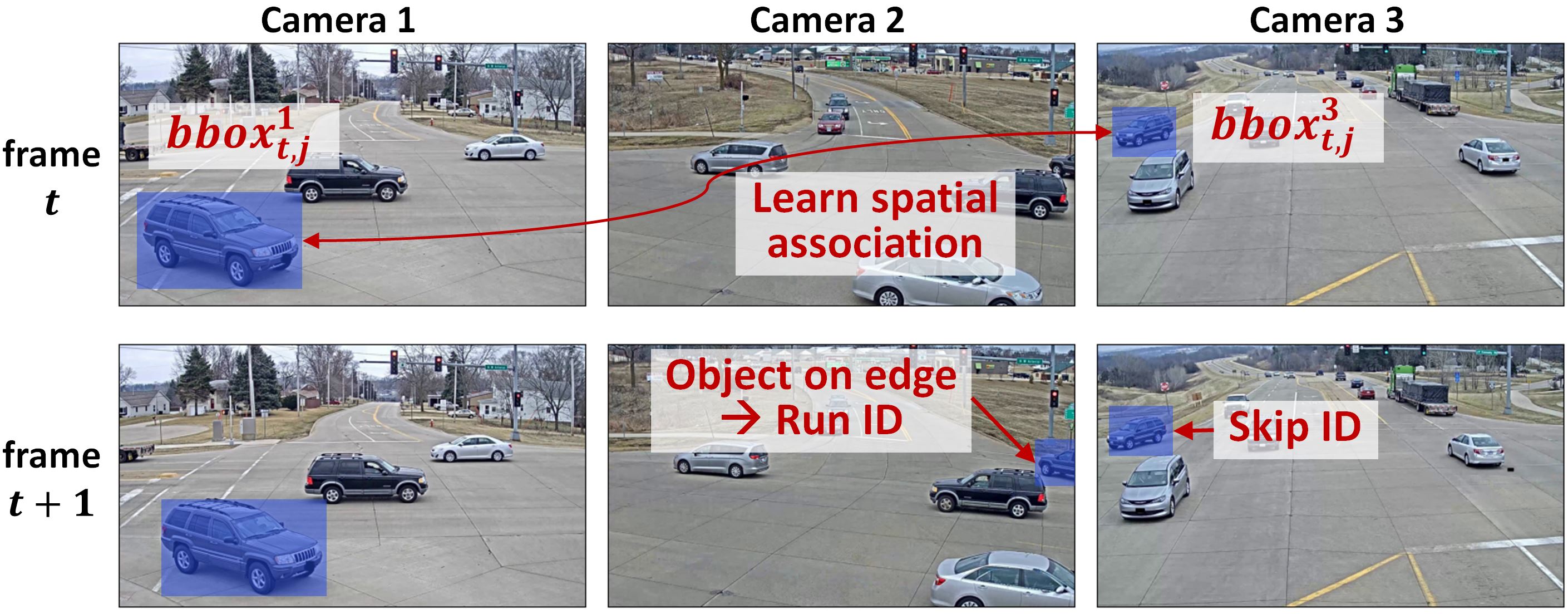}
%    \vspace{-0.1in}
    \caption{Handling newly appearing objects on frame edges.}
    \label{fig:system_edge_exmaple}
%    \vspace{-15pt}
\end{figure}

\subsection{Improving Robustness}~\label{subsec:robustness}

\noindent \textbf{Handling newly appearing objects.} One practical issue that needs to be considered when applying spatial association is how to deal with objects that appear in the FoV for the first time. Figure~\ref{fig:system_edge_exmaple} shows an example. At time $t$ (first row), a target vehicle is found only in Cameras 1 and 3, so the mapping entry is made as $\{bbox_{t,j}^1, N/A, bbox_{t,j'}^3\}$. At time $t+1$ (second row), the target starts to appear in Camera 2. However, if the target is found in Camera 1 and its mapping entry matches the one at time $t$, the target in Camera 2 will not be inspected. To avoid such a case, we skip mapping-based identity matching for objects that appear in the frame for the first time (i.e., we perform an identification task for a vehicle in the blue box in Camera 2 of the second row in Figure~\ref{fig:system_edge_exmaple}) and match its identity based on identification feature matching. Note that we apply for the mapping-based identity matching for other cameras (e.g., Camera 3).

To effectively identify objects when they first appear, we devise a simple and effective heuristic method. Inspired by the observation that an object appears in the camera's frame by moving from out-of-FoV to FoV, we consider the bounding boxes that are newly located at the edge of the frame as potential candidates, and perform the ID feature extraction regardless of the matching mapping entry if no corresponding identification cache is found.

\parjump \noindent 
\revise{\textbf{Handling occlusion.} Depending on a camera's FoV, a target object might be obscured by another moving object. For instance, in a camera with a FoV perpendicular to the road, a vehicle in the front lane could occlude a vehicle in the rear lane. Under such circumstances, the detection model might fail to identify the target object. To handle errors resulting from sudden, short-term occlusions, we develop an interpolation technique that leverages the detection results from the preceding frame in the same camera and/or time-synchronised frames from other cameras. Specifically, during a sudden, short-term occlusion, the target object might be visible up to a certain point in the frame, then abruptly disappear mid-frame. If the object remains visible in other cameras, we can estimate the existence of the occluded object by comparing the current mapping entry with past mapping entries. For example, if an object suddenly disappears in Camera 1, \system{} searches for a prior mapping entry containing the object located in the previous frame of Camera 1 and extracts the position of the object in other cameras. If corresponding bounding boxes are found in all other cameras, \system{} performs object detection and identification on the other cameras. Where occlusion persists for an extended period, we employ periodic cache refreshing (details provided subsequently). It is important to note that such occlusions are rare in practical settings, as objects move at varying speeds and cameras are often installed to monitor the target scene from a high vantage point (e.g., mounted on a traffic light as shown in Figure~\ref{fig:system_edge_exmaple}).}

\parjump \noindent 
\revise{ \textbf{Periodic cache refreshing.} 
To avoid error propagation in our association-based identification (due to occlusion as well as the failure of identification model inference), we limit the maximum number of consecutive skips and perform identification task regardless of a matching mapping entry at the predefined interval (e.g., every 2s). This variable controls the trade-off between efficiency and accuracy.
}

\parjump \noindent \textbf{Time synchronisation.} For spatial association, it is important for all video streams to be time synchronised. To this end, \system{} periodically synchronises the camera clock time using the network time protocol (NTP) periodically and aligns frames based on their timestamp, i.e., two frames are considered time-synchronised if the difference of their timestamps is below the threshold (in the current implementation we set it to 3~ms). Considering that existing CCTV networks are often connected with a gigabit wired connection, \revise{NTP is capable of achieving this value}. Leveraging the synchronized clocks, we match frames across different cameras with the smallest timestamp difference to handle cases where the cameras have different frame rates.
\section{On-Camera Distributed Processing}~\label{sec:distributed}

\subsection{Multi-Camera Dynamic Inspection~\label{subsec:dynamic}}

The key to maximizing the benefit from spatial association is to quickly find the objects that match the query, thereby (a) skipping identification tasks on other cameras from spatial association and (b) skipping identification of objects irrelevant to the query even on the same camera. To this end, we develop a method to dynamically arrange the order of cameras and bounding boxes to be examined.

\parjump \noindent \textbf{Inter-camera dynamic inspection.}
%Consider the example situation illustrated in Figure~\ref{fig:system_approach}, where a target object is captured by Cameras 1 and 2, but not by Camera 4. Assume that all cameras detect the same number of vehicles (e.g. four). If Camera 1 or 2 is inspected first, we can skip the identification operations for Cameras 2 and 4 (i.e., within four identifications). However, if the inspection starts with Camera 4, we need to perform further inspections with Camera 1 or 2 just in case the target object is located out of Camera 4's FoV. Hence, eight identifications are required.
\revise{
The inspection order of the cameras heavily affects the identification efficiency (i.e., the total number of required identifications).
Specifically, we find that searching the cameras which most likely contain the target object first improves the search efficiency.
This is because we can leverage the bounding box location of the identified target object to aggressively skip identification on non-matching bounding boxes in the remaining cameras. 
}
For example, consider a case with three cameras (Cameras 1, 2, and 3). At a given timestamp, assume that all cameras detect the same number of vehicles, e.g., four, and a target object is captured by Cameras 1 and 2. If Camera 1 is inspected first, we can find the query object within four identification and skip the identification operations for Cameras 2 and 3. However, if the inspection starts with Camera 3, we need to perform further inspections with Camera 1 and 2 just in case the target object is located out of Camera 3's FoV. Hence, eight identifications are required.

In addition to efficiency, the inspection order of the cameras also affects the identification accuracy because our approach relies on identification-based target matching from the first camera.
\revise{
Specifically, inspecting the camera where the target identification accuracy is expected to be the highest leads to the highest association accuracy in the remaining cameras.
While the identification accuracy is affected by multiple attributes of the captured object (e.g., the detected object's size, pose, blur), we currently use the bounding box size as the primary indicator; we plan to extend the analysis to other attributes in our future work.
For example, in the case of Figure~\ref{fig:system_spatial_association}, we consider Camera 2 as the first camera to be inspected as the size of the box of the detected target object is the largest, i.e., has the highest probability that it is correctly identified.
}

Considering these two factors (efficiency and accuracy), at each time $t$, we calculate each camera's priority as follows (higher value indicates higher priority)
%we rank the order of the cameras by 
%sorting $\alpha \times \frac{N_{t-1}^i}{N_Q} + (1-\alpha) \times \sum_j^{N_{t-1}^i} c \times size(bbox_{t-1,j}^i)$ for each camera in descending order, 

\begin{equation}
\alpha \times \frac{N_{t-1}^i}{N_Q} + (1-\alpha) \times \sum_j^{N_{t-1}^i} c \times size(bbox_{t-1,j}^i),
\end{equation}

\noindent where $N_{t-1}^i$ is the number of target objects found in $C^i$ at time $t-1$, $N_Q$ is the number of queries, $size()$ is a function that returns the size of the given bounding box, $c$ is a coefficient to normalise the size. $\alpha$ is a variable that determines the weight of resource efficiency and identification accuracy. 

\parjump \noindent \textbf{Intra-camera bounding box dynamic inspection.} 
After object detection in a frame of a single camera, the order of the boxes to be inspected also affects the overall identification performance. 
\revise{
Specifically, inspecting the detected bounding boxes close to the expected location of the target object is more beneficial, as we can skip the identification on the remaining boxes as soon as the target object is identified.
}
For example, in Camera 1 in Figure~\ref{fig:system_approach}, it would be beneficial to start by examining the expected target object (a white vehicle in the third row) rather than starting from the query-irrelevant objects. We order the sequence of boxes to be examined by leveraging temporal association, i.e., sorting 

\begin{equation}
\sum_j min\_dist(bbox_{t,j}^i, B_{t-1}^i)
\end{equation}

%$\sum_j min\_dist(bbox_{t,j}^i, B_{t-1}^i)$ 
\noindent for each bounding box $bbox_{t,j}^i$ in $C^i$ at time $t$ in ascending order, where $B_{t-1}^i$ is a set of bounding boxes of the target objects detected in $C^i$ at time $t-1$ and $min\_dist(bbox, B)$ is a function that returns the minimum distance from $bbox$ to any bounding box in $B$. Note that this only affects the order of the boxes to be examined, but not the tracking result.

\subsection{Multi-Camera Parallel Processing}~\label{subsec:distributing}

The key challenge of running spatio/temporal association on distributed cameras is the long execution time. 
%\revise{\st{While it significantly reduces the total number of ID feature extractions required for identity matching,}} 
The end-to-end execution time may increase if the target objects are not found in the previously inspected cameras due to the sequential execution of the inspection operations. 
%\revise{\st{To optimize the end-to-end execution time,}} 
We apply the following techniques that exploit the resources of the distributed cameras \revise{to prevent this}.

\begin{enumerate}[leftmargin=*]
	\setlength\itemsep{-0.1em}
	\item Given an image, we perform spatial association-irrelevant tasks on the cameras in parallel, i.e., object detection and ID feature extraction of newly appeared objects at the edges of the frame.
	\item If the number of objects in a frame exceeds the pre-defined batch size (e.g., 4), we distribute the identification tasks to nearby cameras and execute them \emph{in parallel}. Such distribution has a beneficial effect on the end-to-end execution time because 1) current AI accelerators do not support parallel execution of AI models~\cite{antonini2019resource}\footnote{Please note that, while Nvidia offers multi-process service (MPS) and multi-instance GPU (MIG) software packages to facilitate model co-running on their GPUS on cloud servers, they are not supported in the Jetson family devices~\cite{jetonAGX,jetonNX} designed for edge AI. The other AI accelerators such as Google Coral TPU and Intel NCS 2 also do not support parallel execution on the accelerator chip.} and 2) the network latency is relatively much shorter since we need to send only the cropped image (e.g., 85$\times$141), not the full-frame image (e.g., 1080p).
\end{enumerate}

Note that batch processing~\cite{crankshaw2017clipper} is widely used to reduce execution time for multiple inferences on a device. To maximize the benefits of workload distribution and batch processing, we profile the execution time with different batch sizes on each camera and network latency with data transmission sizes. Then, we dynamically select the optimal batch size for processing in one camera and the optimal number of bounding boxes for distribution to other cameras.

\revise{
Formally, we define this problem as follows. Given the inspection order determined in \S\ref{subsec:dynamic}, suppose that we are currently processing Camera $i$, where a total of $N$ bounding boxes were detected. We distribute the identification tasks across $K$ cameras to minimize the total execution time as:
}

\begin{equation}
\begin{gathered}
\min_{n^1, n^2, ..., n^K} \max_j \left( TD (C^i, C^j, n^j) + BP (C^j, n^j) \right),\\ 
\text{where} \sum_i n^j=N.
\end{gathered}
\end{equation}

\noindent \revise{where $n^j$ is the number of bounding boxes to distribute to Camera $j$ ($C^j$) to extract the ID features, $TD(C^i, C^j, n^j)$ is a function that returns the network transmission delay to distribute $n^j$ cropped images from $C^i$ to $C^j$ (note that $TD(C^i, C^i, n^i)$ is zero as no transmission is required), and $BP(C^j, n^j)$ is the batched processing latency the identification model on $C^j$. $TD(C^i, C^j, n^j)$ and $BP(C^j, n^j)$ vary for each $C^j$ depending on its processing capability and network bandwidth; we predict $TD(C^i, C^j, n^j)$ and $BP(C^j, n^j)$ as follows.
}

\revise{
First, transmission delay for the workload distribution $TD(C^i, C^j, n^j)$ is calculated as 
}

\begin{equation}
TD (C^i, C^j, n^j) = H\cdot W \cdot n^j/ BW_i^j,
\end{equation}

\revise{
\noindent where $H,W$ is the height and width of the image crop (resized to the input size of the identification model) and $BW_i^j$ is the network bandwidth between $C^i$ and $C_j$ ($BW_i^i$ is set as $\infty$). For each network transmission event between $C^i$ and $C_j$, we estimate $BW_i^j$ by the transmitted data size divided by the transmission latency (similar to \cite{laskaridis2020spinn}), and update it with Exponential Weighted Moving Average (EWMA) filtering for future prediction.
In our current implementation, the cameras are connected with a Gigabit wired connection similar to \cite{zeng2020distream}, and the distribution latency is negligible compared to the identification model inference latency (e.g., $\approx$0.3~ms to transmit a 128$\times$128 cropped image over 1~Gbps connection, whereas single identification takes $>$100~ms as shown in Table~\ref{tab:motiv_benchmark}).
}

\revise{
Next, identification latency with batched processing $BP(C^i, n^i)$ is calculated as
}
\begin{equation}
BP(C^i, n^i) = \lceil n^i/n_{batch}^i \rceil \times T(C^i, n_{batch}^i),
\end{equation}
\noindent \revise{
where $n_{batch}^i$ is the batch size on $C^i$, and $T(C^i, n_{batch}^i)$ is the identification model latency on $C^i$ with batch size $n_{batch}^i$. 
$n_{batch}^i$ is determined at the offline stage by running the identification model on each $C^i$ with different batch sizes and determining the one that maximizes the throughput. At runtime, we update $T(C^i, n_{batch}^i)$ upon each inference using the EWMA to account for dynamic resource fluctuation (e.g., due to thermal throttling) similar to \cite{jeong2022band}.
}

\section{Evaluation}
\label{sec:evaluation}

\subsection{Experimental Setup~\label{subsec:eval_setup}}

\begin{figure}[t]
    \centering
	\minipage{1.0\columnwidth}
    \includegraphics[width=1\textwidth]{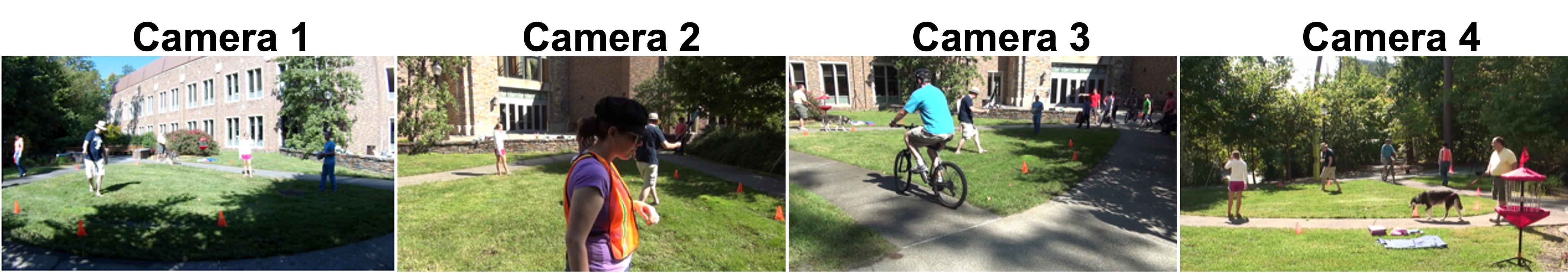}
	\caption{Snapshots of CAMPUS-Garden1~\cite{xu2016campus} dataset.}
    \label{fig:6-eval-campus-topology}
	\endminipage\hfill
\end{figure}

\textbf{Datasets.} We use three real-world overlapping camera datasets for the evaluation: CityFlowV2~\cite{naphade20215aicity}, CAMPUS ~\cite{xu2016campus}, and MMMPTRACK~\cite{han2023mmptrack} \revise{to ensure a fair comparison with baseline methods and to enable an in-depth study of the impact of various system parameter values.} When spatial association is used, we use the pre-generated mapping entries learned with 10\% of the data in the dataset.
\begin{itemize}[leftmargin=*]
	\item {\bf CityFlowV2}~\cite{naphade20215aicity} consists of video streams from five heterogeneous overlapping cameras at a road intersection. The cameras are located to cover the intersection from different sides of the road (Figure~\ref{fig:motiv_coverage_topology}). 4 videos are recorded at 1080p@10fps and 1 video is recorded at 720p@10fps with a fisheye lens. Each video stream is $\approx$3 minutes long and the ground truth data contains 95 unique vehicles.
 
	\item {\bf CAMPUS }~\cite{xu2016campus} consists of overlapping video streams recorded in four different scenes. We use the \textit{Garden1} scene, which consists of 4$\times$ 1080@30fps videos capturing a garden and its perimeter (Figure~\ref{fig:6-eval-campus-topology}). We resized the images to 720p as they show comparable object detection performance to the original 1080p at a lower cost. Each video is $\approx$100s and the ground truth contains 16 unique individuals. Since the dataset provides inaccurate ground truth labels and bounding boxes, we manually regenerated the ground truth for three targets (id 0, 2, 9).
 
    \item {\bf MMPTRACK~\cite{han2023mmptrack}} \revise{is composed of overlapping video streams recorded from 5 different scenes: \emph{cafe shop, industry safety, office, lobby, and retail}. In total, there are 23 scene samples (3-8 samples per scene), and each sample is composed of 4-6 overlapping video streams capturing 6-8 people. Each video stream is 360p@15fps and $\approx$400 seconds (in total 133k frames = 8,800 seconds). We use this dataset to evaluate the robustness of \system{} in \S\ref{subsec:eval-robustness}}
\end{itemize}

\parjump
\noindent \textbf{Queries.} For queries, we randomly chose ten vehicles for CityFlowV2, three people for CAMPUS, and two people for MMPTRACK. In the in-depth analysis, we also examine performance with different numbers of queries.

\parjump
\noindent \textbf{Object detection and identification models.} For object detection, we use YOLO-v5~\cite{yolov5}. For vehicle identification in CityFlowV2, we use the ResNet-101-based model~\cite{luo2021empirical} trained on the CityFlowV2-ReID dataset~\cite{naphade20215aicity}. For person identification in CAMPUS and MMPTRACK, we trained the ResNet-50-based model using the dataset. Note that the performance of the re-id model is not the focus of this work and different models can be used. All models are implemented in PyTorch 1.7.1.

\parjump
\noindent \textbf{Metrics.} To measure system resource costs, we evaluate the end-to-end latency and the number of identification model inferences. To measure tracking quality, we use two metrics that are widely used in multi-object tracking~\cite{bernardin2006multiple}: Multiple Object Tracking Precision (MOTP) and Multiple Object Tracking Accuracy (MOTA).
\begin{itemize}[leftmargin=*]
	\item \textbf{End-to-end latency} is the total latency for generating multi-camera, multi-target tracking results. Note that the latency includes all the operations required for the system, i.e., image acquisition, model inference, uploading the cropped images to other cameras, and cross-camera communication time. 
	\item \textbf{Number of IDs} is defined as the total number of identification model inferences required across all cameras for each timestamp.
	\item \textbf{MOTP} quantifies how precisely the tracker estimates object positions. It is defined as $\frac{\sum_{t,i} d_{t,i}}{\sum_{t} c_t}$, where $c_t$ is the number of matches in frame $t$ and $d_{t.i}$ is the overlap of the bounding box (IoU) of target $i$ with the ground truth. For each frame, we compute the MOTP for each camera separately and report its average.
	\item \textbf{MOTA} measures the overall accuracy of both the detector and the tracker. We define it as $1 - \frac{\sum_{t} ( FN_t + FP_t + MM_t )}{\sum_{t} T_t}$, where $t$ is the frame index and $T_t$ is the number of target objects in frame $t$. FN, FP, and MM represent false-negative, false-positive and miss-match errors, respectively. Similarly, we calculate the average MOTA across multiple cameras and report their average.
\end{itemize}

\parjump
\noindent \textbf{Baselines.} We evaluate \system{} against the following state-of-the-art methods. The baselines perform all model operations on the camera where the corresponding image frame is generated.

\begin{itemize}[leftmargin=*]
	\item \textbf{Conv-Track} is the conventional pipeline of multi-camera, multi-target tracking (e.g., \cite{liu2021city}), as shown in Figure~\ref{fig:2-background-pipeline}. It identifies the query object on each camera separately and aggregates the identification results across multiple cameras.
	\item \textbf{Spatula-Track} adopts the camera-wise filtering approach proposed in Spatula~\cite{jain2020spatula} for object tracking. For each timestamp, it first filters out the cameras that do not contain the target objects and then performs the Conv-Track pipeline for the selected cameras. We use ground truth labels for correlation learning and camera filtering, assuming the ideal operation of Spatula~\cite{jain2020spatula}.
	\item \textbf{CrossRoI-Track} adopts the RoI-wise filtering approach proposed in CrossRoI~\cite{guo2021crossroi} for tracking. Offline, it learns the minimum-sized RoI that contains all objects at least once over deployed cameras. At runtime, it performs the Conv-Track pipeline only for the masked RoI areas. We use the ground truth labels for the optimal training of the RoI mask, assuming the ideal operation of CrossRoI~\cite{guo2021crossroi}.
\end{itemize}

\parjump
\noindent \textbf{Hardware.} For the hardware of smart cameras, we considered two platforms, Nvidia Jetson AGX and Jetson NX. Jetson AGX hosts an 8-core Nvidia Carmel Arm, a 512-core Nvidia VoltaTM GPU with 64 Tensor Cores and 32 GB of memory. Jetson NX hosts a 6-core Nvidia Carmel Arm, a Volta GPU with 384 NVIDIA CUDA cores and 48 Tensors, and 8 GB of memory. We prototyped \system{} on these platforms and measured performance; we used Jetson AGX for the CityFlowV2 dataset and the MMMPTRACK dataset, and Jetson NX for the CAMPUS dataset. For the network configurations, we connected the Jetson devices with a Gigabit wired connection, which is commonly used for existing CCTV networks. It is important to note that, while we used the offline data traces from three datasets for the repetitive and comprehensive analysis, we implemented the end-to-end, distributed architecture of \system{} on top of multiple Jetson devices and evaluated the resource metrics by monitoring the resource cost at runtime. 

\begin{figure*}[t]
    \centering
	\minipage{0.5\textwidth}
    \begin{subfigure}[b]{0.49\textwidth}
        \centering
        \includegraphics[width=\textwidth]{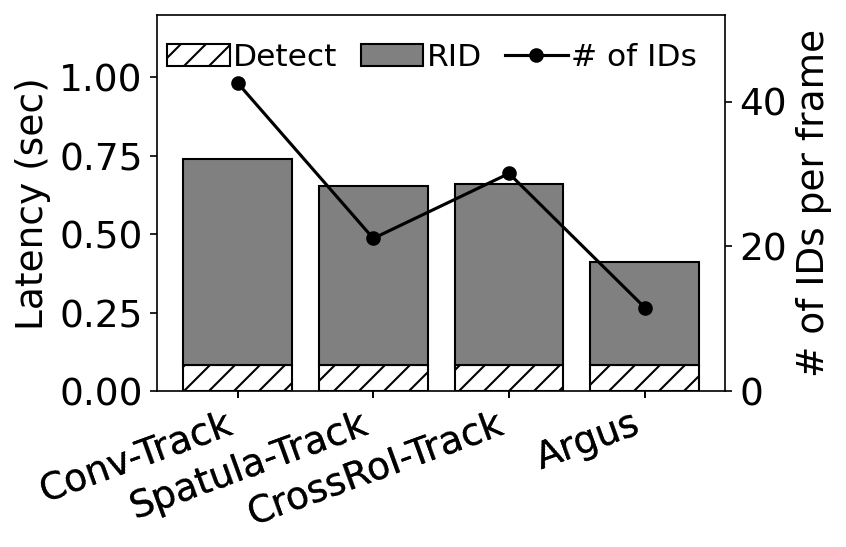}
    	\vspace{-0.2in}
    	\caption{Resource cost.}
    	\label{fig:eval_overall_aicity_resource}
    \end{subfigure}
    \begin{subfigure}[b]{0.49\textwidth}
		\centering
        \includegraphics[width=\textwidth]{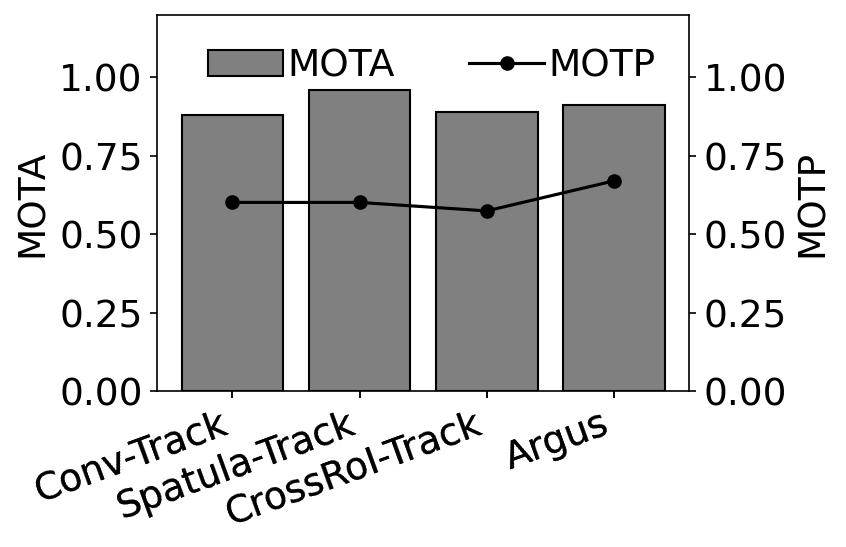}
	    \vspace{-0.2in}
    \caption{Tracking quality.}
    \label{fig:eval_overall_aicity_quality}
    \end{subfigure}
    %\vspace{-0.15in}
    \caption{Overall performance on CityFlowV2.}
    \label{fig:eval_overall_aicity}
	\endminipage
	\minipage{0.5\textwidth}
    \begin{subfigure}[b]{0.49\textwidth}
        \centering
        \includegraphics[width=\textwidth]{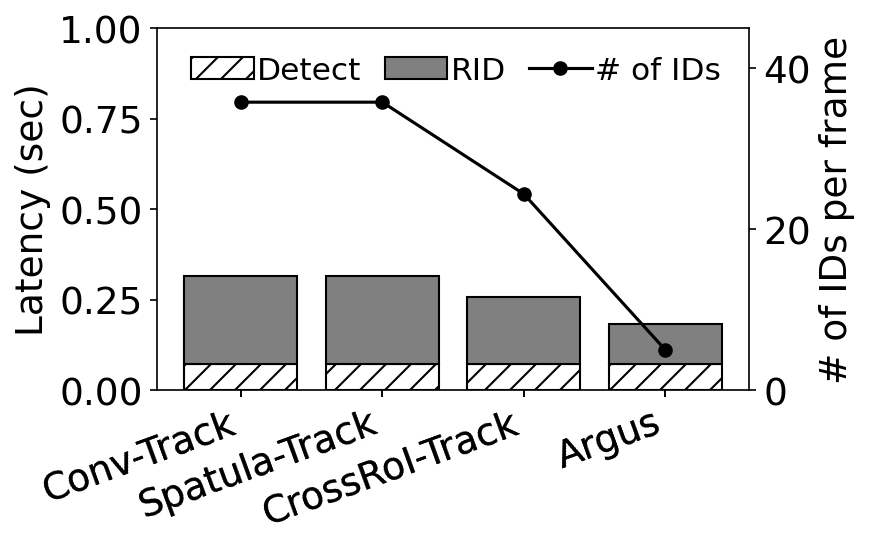}
    	\vspace{-0.2in}
    	\caption{Resource cost.}
    	\label{fig:eval_overall_campus_resource}
    \end{subfigure}
    \begin{subfigure}[b]{0.49\textwidth}
    	\centering
        \includegraphics[width=\textwidth]{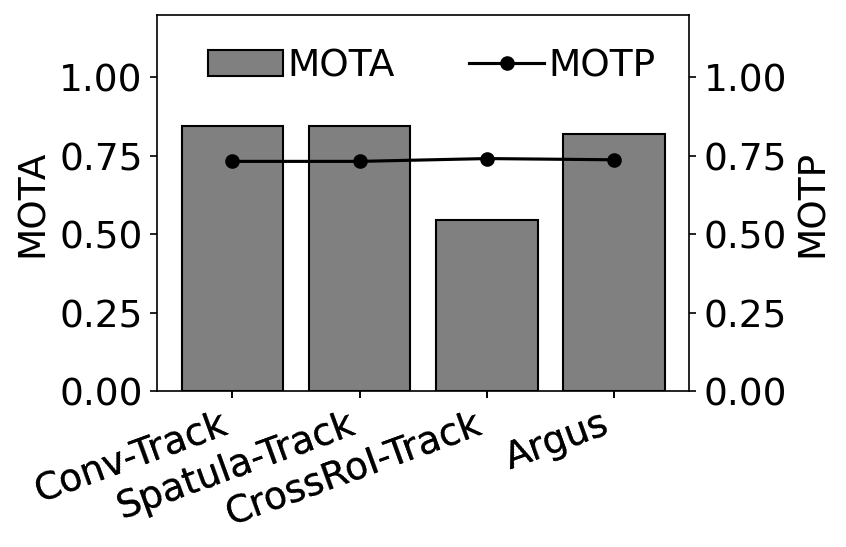}
    	\vspace{-0.2in}
    	\caption{Tracking quality.}
    	\label{fig:eval_overall_campus_quality}
    \end{subfigure}
    %\vspace{-0.15in}
    \caption{Overall performance on CAMPUS.}
    \label{fig:eval_overall_campus}
	\endminipage
\vspace{-10pt}
\end{figure*}

\subsection{Overall Performance}~\label{subsec:overall_performance}

Figures~\ref{fig:eval_overall_aicity} and \ref{fig:eval_overall_campus} show overall performance on CityFlowV2 and CAMPUS respectively. In Figures~\ref{fig:eval_overall_aicity_resource} and \ref{fig:eval_overall_campus_resource}, the bar chart represents the average end-to-end latency (Detect: object detection latency, ID: identification latency) and the line chart shows the average number of IDs. In Figures~\ref{fig:eval_overall_aicity_quality} and \ref{fig:eval_overall_campus_quality}, the bar and line charts represent the average MOTA and MOTP respectively.

\subsubsection{Resource Efficiency}

Overall, \system{} achieves significant resource savings by adopting the spatio-temporal association and workload distribution, while not compromising the tracking quality. We first examine the resource saving of \system{}. In CityFlowV2 where five cameras are involved, Figure~\ref{fig:eval_overall_aicity_resource} shows that the average number of IDs decreases from 42.6 (Conv-Track ) to 21.1 (Spatula-Track), 30.1 (CrossRoI-Track) and 11.6 (\system{}). The end-to-end latency also decreases from 740 ms to 650 ms, 660 ms and 410 ms, respectively; \system{} is 1.8$\times$, 1.59$\times$ and 1.61$\times$ faster than Conv-Track, Spatula-Track and CrossRoI-Track, respectively, which are the state-of-the-art multi-camera tracking solutions. We find several interesting observations. First, the latency does not decrease proportionally to the number of IDs because all baselines need to commonly perform object detection in every frame. However, even when object detection is taken into account, \system{} significantly decreases the end-to-end latency by 49\% by reducing the number of IDs by 73\%, compared to Conv-Track. Second, both Spatula-Track and CrossRoI-track significantly reduce the number of IDs by selectively using cameras and RoI areas, respectively. However, the reduction in end-to-end latency is not significant (about 10\%). This is because the latency is tied to the longest execution time of all cameras due to the lack of distributed processing capability.

Figure~\ref{fig:eval_overall_campus_resource} compares the resource costs for the CAMPUS dataset. The results show a similar pattern to the CityFlowV2 dataset, but the saving ratio of \system{} is much higher. \system{} reduces the average number of IDs by 7.13$\times$ (35.8 to 5.0) compared to Conv-Track and Spatula-Track and 4.86$\times$ (24.4 to 5.0) compared to CrossRoI-Track. The end-to-end latency also decreases by 1.72$\times$ (from 310 ms to 180 ms) and 1.43$\times$ (from 258 ms to 180 ms), respectively. The larger saving is mainly because the moving speed of the target objects (here, people in the CAMPUS dataset) is relatively slow, compared to vehicles in the CityFlowV2 dataset. Therefore, there are fewer newly appearing objects (at the edge) and most of the identification tasks can be done by spatial and temporal association matching. Interestingly, Spatula-Track shows the same performance as Conv-Track, which is different from the CityFlowV2 case. This is because all target people are captured by all four cameras all the time and thus cameras are not filtered out. CrossRoI-Track reduces both latency and the number of IDs compared to Conv-Track, but its efficiency is still lower than \system{}; for CrossRoI-Track, the number of IDs and latency are 24.9 and 419 ms, respectively.

\subsubsection{Tracking Quality}

We investigate how spatial and temporal association-aware identification affects tracking quality. Figure~\ref{fig:eval_overall_aicity_quality} and Figure~\ref{fig:eval_overall_campus_quality} show the MOTP and MOTA on CityFlowV2 and CAMPUS, respectively. Overall, \system{} achieves comparable tracking quality, even with significant resource savings. Interestingly, in CityFlowV2, \system{} increases both MOTA and MOTP compared to Conv-Track; MOTA increases from 0.88 to 0.91 and MOTP increases from 0.60 to 0.67. This is because several small cropped vehicles are identified by associating with their position from other cameras, which failed to be identified by matching their appearance features from the identification model in the baselines. In CAMPUS, \system{} shows almost the same tracking quality as Conv-Track, but MOTA drops slightly from 0.85 (Conv-Track) to 0.82. There were some cases where a target person was suddenly occluded by another person in some cameras. However, \system{} identifies the occluded person in the next frame using the robustness techniques~\S\ref{subsec:robustness}, thereby being able to minimize the error.

We investigate the benefit of cross-camera collaboration in more detail. In CAMPUS, Spatula also increases the tracking accuracy (MOTA) compared to Conv-Track by filtering out query-irrelevant cameras, thereby avoiding false-positive identifications. However, in CAMPUS, its quality is identical to Conv-Track as no cameras are filtered out. Unlike Spatula-Track, CrossRoI degrades the tracking quality on both MOTA and MOTP; for example, in CAMPUS, CrossRoI shows 0.55 of MOTA, while other baselines including \system{} show 0.85 of MOTA. This is because, for object detection and identification, CrossRoI uses the smallest RoI across all cameras in which the target objects appear at least once. Therefore, these tasks sometimes fail due to the small size of the objects in the generated RoI areas.

\subsection{Performance Breakdown: Benefit of On-Camera Distributed Processing}

\begin{figure}[t!]
    \centering
    \begin{subfigure}[b]{0.23\textwidth}
        \centering
        \includegraphics[width=\textwidth]{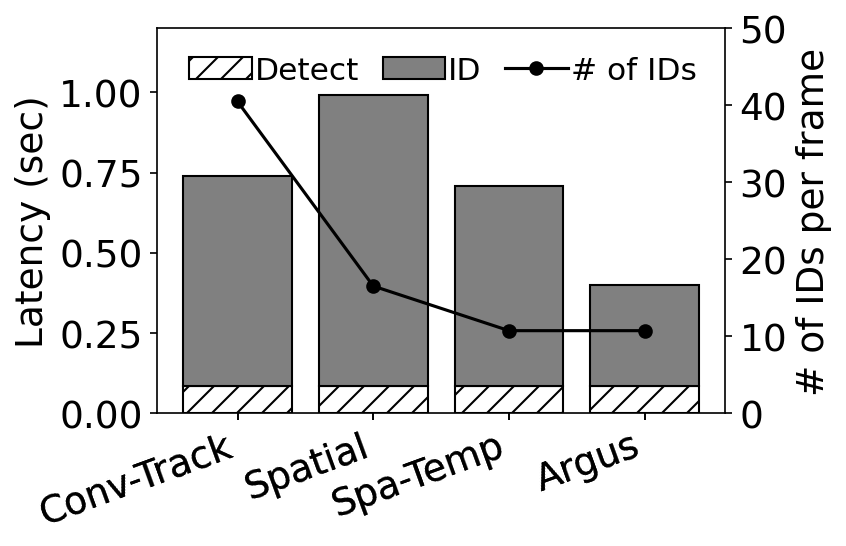}
    	\vspace{-0.2in}
    	\caption{Resource cost.}
    	\label{fig:eval_breakdown_aicity_resource}
    \end{subfigure}
    \begin{subfigure}[b]{0.23\textwidth}
		\centering
        \includegraphics[width=\textwidth]{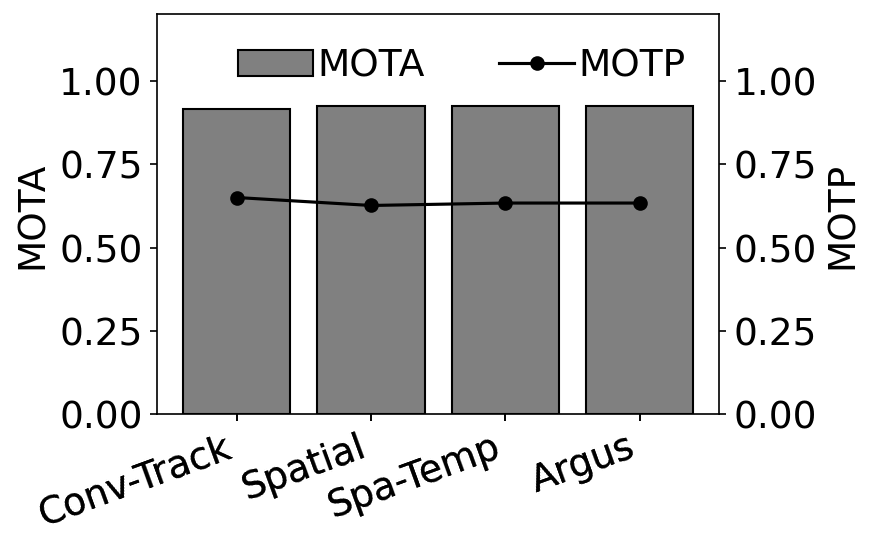}
	    \vspace{-0.2in}
    \caption{Tracking quality.}
    \label{fig:eval_breakdown_aicity_quality}
    \end{subfigure}
    %\vspace{-0.15in}
    \caption{Performance breakdown on CityFlowV2.}
    \label{fig:eval_breakwdown_aicity}
\vspace{-10pt}
\end{figure}

We developed two variants of \system{}, namely \textit{Spatial} and \textit{Spa-Temp}, in which we apply each enhancement to Conv-Track in turn. For identification optimization, Spatial uses spatial association and Spa-Temp uses the spatio/temporal association. Both of them have the capability of dynamic inspection of cameras and bounding boxes (\S\ref{subsec:dynamic}), but do not have distributed processing (\S\ref{subsec:distributing}).

Figure~\ref{fig:eval_breakdown_aicity_resource} shows the resource cost in CityFlowV2. The spatial association reduces the number of IDs from 40.5 (Conv-Track) to 16.5 (Spatial) and the temporal association further decreases to 11.6 (Spa-Temp). These results show that our spatial and temporal association techniques make a significant contribution to overall resource savings. Interestingly, despite the reduction in the number of IDs by Spatial, the latency increases from 854 ms (Conv-Track) to 992 ms (Spatial) due to the sequential operations on the cameras. However, we observe that the temporal association and workload distribution successfully reduce the latency in turn, to 624 ms and 317 ms, respectively. Figure~\ref{fig:eval_breakdown_aicity_quality} shows the tracking quality of CityFlowV2. It confirms again that both spatial and temporal associations (and their mapping-based identity matching) do not compromise the tracking quality. We omit the result of CAMPUS as we also observe a similar trend.

\subsection{Impact of Number of Queries}

\begin{figure}[t]
  \centering
  \begin{minipage}[b]{0.45\textwidth}
    \centering
    \begin{subfigure}[b]{0.49\textwidth}
      \centering
      \includegraphics[width=\textwidth]{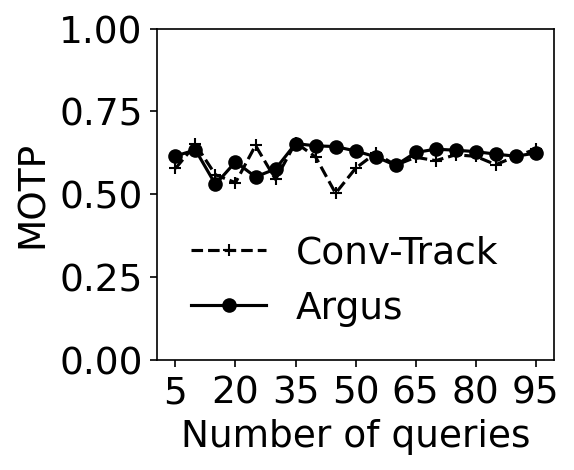}
      \caption{MOTP.}
      \label{fig:eval_nqueries_aicity_motp}
    \end{subfigure}
    % \hfill
    \begin{subfigure}[b]{0.49\textwidth}
      \centering
      \includegraphics[width=\textwidth]{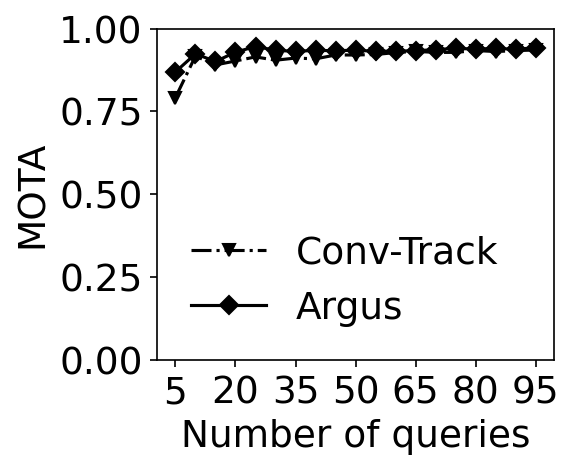}
      \caption{MOTA.}
      \label{fig:eval_nqueries_aicity_mota}
    \end{subfigure}
    \caption{Impact of the number of queries on tracking quality (CityFlowV2).}
    \label{fig:eval_nqueries_campus_quality}
  \end{minipage}
  \hfill
  \begin{minipage}[b]{0.45\textwidth}
    \centering
    \begin{subfigure}[b]{0.49\textwidth}
      \centering
      \includegraphics[width=\textwidth]{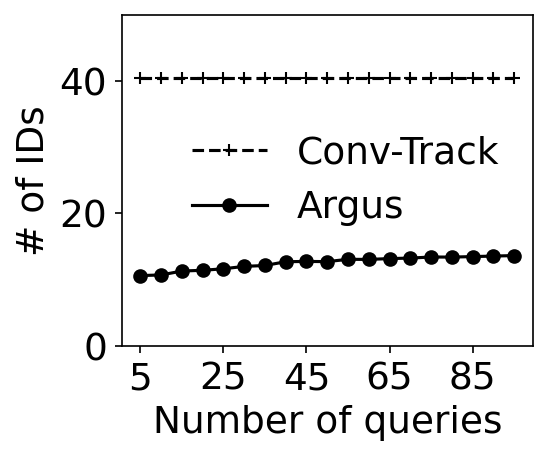}
      \caption{Number of IDs.}
      \label{fig:eval_nqueries_aicity_nreid}
    \end{subfigure}
    \hfill
    \begin{subfigure}[b]{0.49\textwidth}
      \centering
      \includegraphics[width=\textwidth]{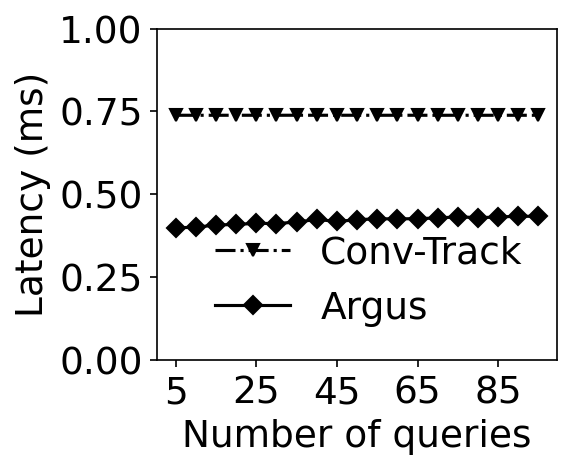}
      \caption{End-to-end latency.}
      \label{fig:eval_nqueries_aicity_latency}
    \end{subfigure}
    \caption{Impact of the number of queries on resource cost (CityFlowV2).}
    \label{fig:eval_nqueries_aicity_resource}
  \end{minipage}
\end{figure}

We investigate the impact of the number of queries on system performance. For CityFlowV2, we vary the number of queries from 5 to 95 with an interval of 5. For each number of queries, we randomly select three sets of queries (except the entire set) and report their average performance. Figures~\ref{fig:eval_nqueries_aicity_motp} and \ref{fig:eval_nqueries_aicity_mota} show MOTP and MOTA for CityFlowV2, respectively; we observe a similar trend in CAMPUS. 
\revise{
For Conv-Track, both the MOTP and MOTA are not significantly affected by the number of queries. This is because Conv-Track runs the identification model on all the detected objects across all cameras regardless of the number of queries; the identification matching accuracy with the query images does not vary with the number of queries as they are randomly selected and averaged. 
\system{} also shows comparable accuracy with Conv-Track (with significantly reduced number of ID operations as shown in Figure~\ref{fig:eval_nqueries_aicity_resource}, showing that it effectively reduces the identification workload without accuracy drop.
Even with a large number of matching attempts with other cameras and queries, \system{} identifies the objects accurately.
}

Figure~\ref{fig:eval_nqueries_aicity_resource} shows how the number of IDs and the end-to-end latency change depending on the number of queries. The number of IDs and the latency in Conv-Track do not change because all objects in the frame must be examined regardless of the number of queries. In contrast to Conv-Track, Figure~\ref{fig:eval_nqueries_aicity_nreid} shows that the number of IDs in \system{} increases with the number of queries. This is because, at each time, if all target objects are not found on the previously inspected cameras, \system{} has to perform the identification operation for all detected objects (which are not filtered out of the spatio/temporal association). This probability increases when the number of queries is large, thereby increasing the number of IDs. However, it saturates when the number of queries is around 50 and, more importantly, it is still much lower than Conv-Track.

Figure~\ref{fig:eval_nqueries_aicity_latency} also shows an interesting result. While the number of IDs of \system{} increases by 29\% from 10.6 to 13.7 when the number of queries is 5 and 95, respectively, the increase in latency is much lower, i.e., by 8\% from 399 ms to 433 ms. If we exclude the latency for object detection for the analysis, the execution time for identification increases by only 10\%, from 315 ms to 349 ms. This result shows the benefit of the \system{}'s distribution of the identification operations to other cameras.

\begin{figure}[t]
  \centering
  \begin{minipage}[b]{0.45\textwidth}
    \centering
    \begin{subfigure}[b]{0.49\textwidth}
      \centering
      \includegraphics[width=\textwidth]{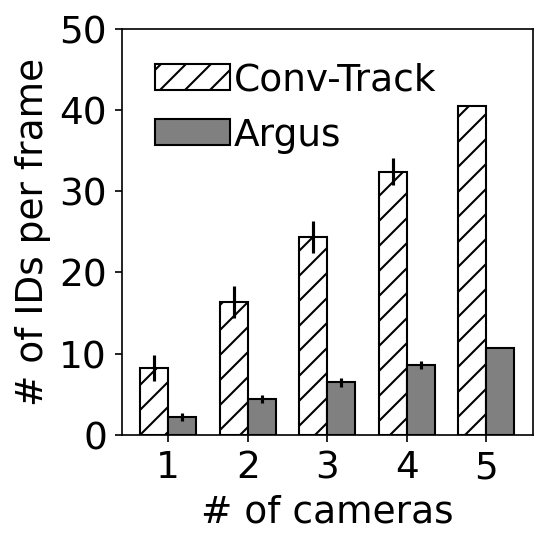}
      \caption{CityFlowV2.}
      \label{fig:eval_ncameras_aicity_nreid}
    \end{subfigure}
    % \hfill
    \begin{subfigure}[b]{0.49\textwidth}
      \centering
      \includegraphics[width=\textwidth]{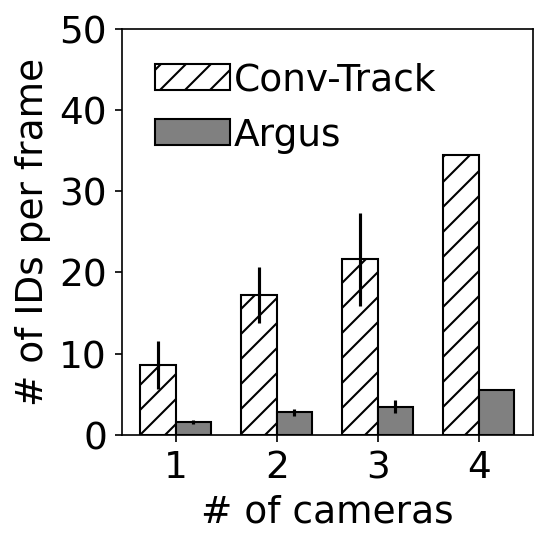}
      \caption{CAMPUS.}
      \label{fig:eval_ncameras_campus_nreid}
    \end{subfigure}
    \caption{Impact of number of cameras on number of IDs.}
    \label{fig:eval_ncameras_nreid}
  \end{minipage}
  \hfill
  \begin{minipage}[b]{0.45\textwidth}
    \centering
    \begin{subfigure}[b]{0.49\textwidth}
      \centering
      \includegraphics[width=\textwidth]{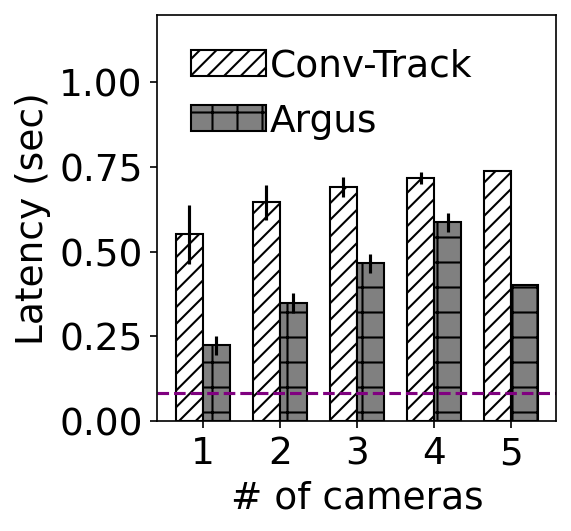}
      \caption{CityFlowV2.}
      \label{fig:eval_ncameras_aicity_latency}
    \end{subfigure}
    \hfill
    \begin{subfigure}[b]{0.49\textwidth}
      \centering
      \includegraphics[width=\textwidth]{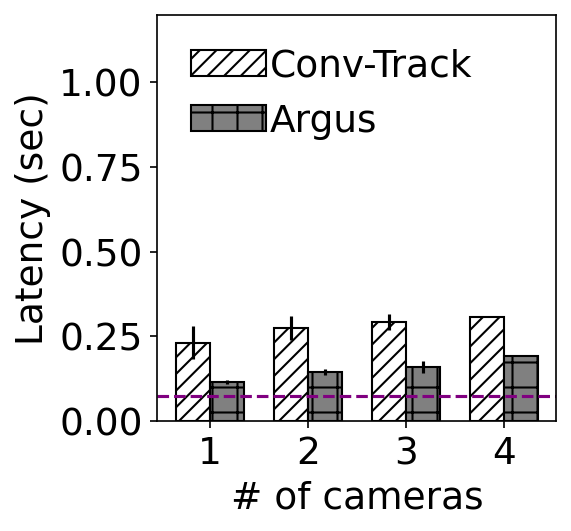}
      \caption{CAMPUS.}
      \label{fig:eval_ncameras_campus_latency}
    \end{subfigure}
    \caption{Impact of the number of cameras on latency; purple line is the execution time of the object detection.}
    \label{fig:eval_ncameras_latency}
  \end{minipage}
\end{figure}
  
\subsection{Impact of Number of Cameras}

We examine the impact of the number of cameras on resource saving. We consider all possible combinations and report their average performance; for example, in the case of three cameras in the CityFlowV2, we report the average result for all 10 (=$_5C_3$) combinations. In this subsection, we do not report the tracking quality results because it is not fair to compare tracking quality for different number and topology of cameras.

Figure~\ref{fig:eval_ncameras_nreid} shows the total number of IDs for Conv-Track and \system{}. As expected, the number of IDs required for both Conv-Track and \system{} increases when more cameras are used. As shown in Figure~\ref{fig:eval_ncameras_aicity_nreid}, in Conv-Track the number of IDs increases from 8.2 (1 camera) to 42.5 (5 cameras) in CityFlowV2, i.e., by 418\%. Similarly, in \system{}, it increases from 2.2 to 11.6, i.e., by 423\%. Figure~\ref{fig:eval_ncameras_campus_nreid} also shows that in CAMPUS, the total number of IDs increases by 316\%, from 8.6 (1 camera) to 35.8 (4 cameras) in Conv-Track, while in \system{}, it increases from 1.6 to 5.0, i.e., by 213\%. However, interestingly, \system{} shows a much lower standard deviation across different combinations. This is because the number of IDs in Conv-Track is proportional to the number of objects in a frame and is therefore affected by the camera's FoV, i.e., how many objects are captured. In contrast, the number of IDs in \system{} is determined by the spatial association of the target objects, and is therefore more dependent on the number of queries.

\iffalse
\begin{figure}[t!]
    \centering
    \begin{subfigure}[b]{0.24\textwidth}
            \centering
            \includegraphics[width=\textwidth]{figures/eval/ncamera_aicity_latency.png}
%    \vspace{-0.1in}
    \caption{CityFlowV2.}
    \label{fig:eval_ncameras_aicity_latency}
    \end{subfigure}
    \begin{subfigure}[b]{0.24\textwidth}
            \centering
            \includegraphics[width=\textwidth]{figures/eval/ncamera_campus_latency.png}
%    \vspace{-0.1in}
    \caption{CAMPUS.}
    \label{fig:eval_ncameras_campus_latency}
    \end{subfigure}
%    \vspace{-0.15in}
    \caption{Impact of the number of cameras on latency; purple dotted line is the execution time of the object detection.}
    \label{fig:eval_ncameras_latency}
%\vspace{-15pt}
\end{figure}
\fi

We further investigate the impact of the number of cameras on end-to-end latency. Figure~\ref{fig:eval_ncameras_aicity_latency} and \ref{fig:eval_ncameras_campus_latency} show the latency in the CityFlowV2 and CAMPUS datasets, respectively. While Conv-Track performs the identification operations on each camera individually, the latency increases as the number of cameras increases. This is because Conv-Track's latency is tied to the maximum latency across all cameras. \system{} also increases latency for both CityFlowV2 and CAMPUS when more cameras are involved, as the waiting time for the entry matching of previously inspected cameras also increases. Nevertheless, the latency of \system{} in both Figures~\ref{fig:eval_ncameras_aicity_latency} and \ref{fig:eval_ncameras_campus_latency} is still much lower than that of Conv-Track. We observe an interesting case in the CityFlowV2 dataset. The latency of \system{} decreases from 587 ms to 400 ms when the number of cameras increases from 4 to 5, even though the number of IDs increases from 8.6 to 11.6. We conjecture that more cameras provide more opportunities for the parallel processing of IDs across distributed cameras, and the benefit becomes apparent when all five cameras are involved.

\subsection{Impact of Inspection Order}

\begin{figure}[t]
    \centering
    %\begin{subfigure}[b]{0.3125\textwidth}
    \begin{subfigure}[b]{0.24\textwidth}
            \centering
            \includegraphics[width=\textwidth]{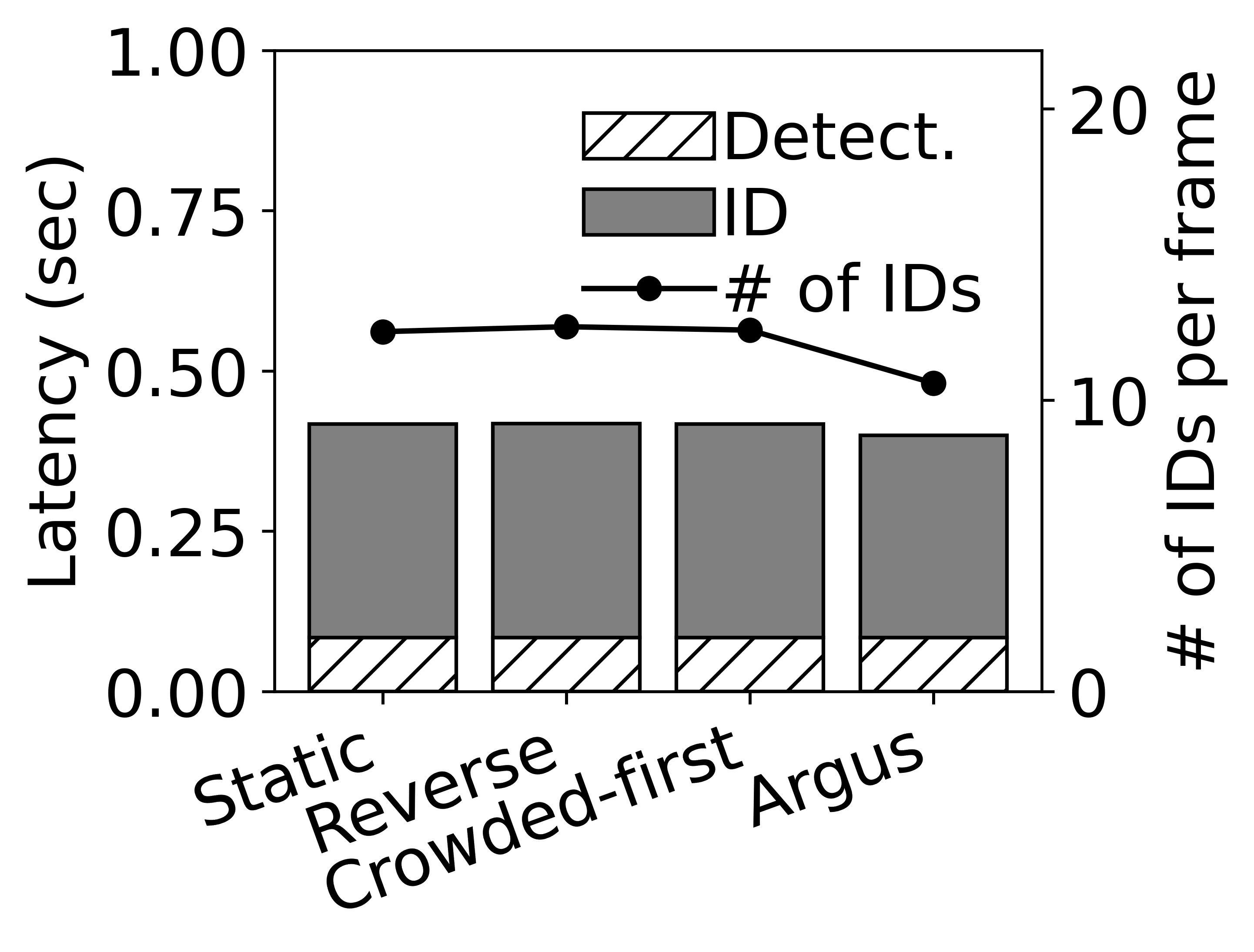}
%    \vspace{-0.1in}
    \caption{CityFlowV2.}
    \label{fig:eval_order_aicity}
    \end{subfigure}
    %\begin{subfigure}[b]{0.3125\textwidth}
    \begin{subfigure}[b]{0.24\textwidth}
            \centering
            \includegraphics[width=\textwidth]{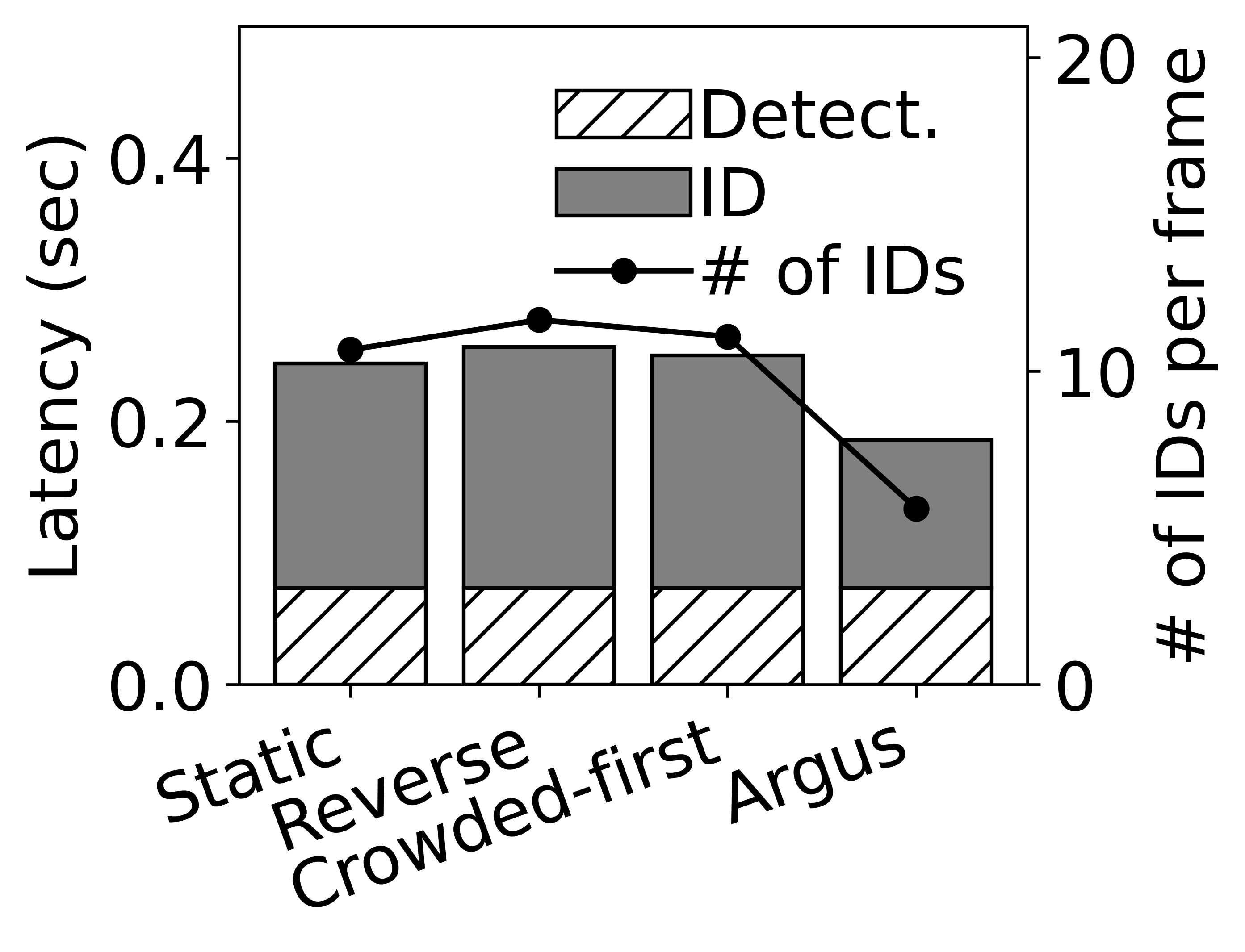}
%    \vspace{-0.1in}
    \caption{CAMPUS.}
    \label{fig:eval_order_campus}
    \end{subfigure}
    \vspace{-0.1in}
    \caption{\revise{Impact of inspection order.}}
    \label{fig:eval_oder}
%\vspace{-10pt}
\end{figure}

\revise{We investigate the impact of the inspection order on system performance. For the study, we developed three variants of \system{}, namely \textit{Static}, \textit{Reverse}, and \textit{Crowded-first}. All of them are built upon the original \system{} system. The Static and Reverse variants inspect the cameras and bounding boxes in a predefined static order and in the reverse order of \system{}, respectively; Reverse is used to establish the performance lower bound and validate our design choice. The Crowded-first variant, inspired by REV~\cite{xu2022rev}, inspects the cameras in a descending order based on the number of bounding boxes. The underlying rationale is that cameras with more bounding boxes are more likely to capture objects of interest. For the bounding box inspection order, we employed the same order as used in Static.} 

Figures~\ref{fig:eval_order_aicity} and \ref{fig:eval_order_campus} show the latency and the total number of IDs in CityFlowV2 and CAMPUS, respectively. We omit the result of MOTP and MOTA as the differences were marginal. The results validate our design choice. In both datasets, \system{} shows shorter latency and fewer identification operations. \revise{As expected, Static and Crowded-first show better performance than Reverse, though their effect is still lower than \system{}. This is primarily due to a lack of considerations for the relevance of target objects in a scene.} This advantage is more evident in CAMPUS. The number of IDs of Reverse is 11.6, while \system{}'s number is 5.0. Similarly, the latency decreases from 245 ms to 186 ms. This is mainly because the target people mostly remain in one of the cameras during the video stream. Therefore, \system{} is capable of reducing the number of IDs by initiating the inspection with potential target objects. 

\begin{figure*}[t]
    \centering
	\minipage{1\textwidth}
    \centering
    \begin{subfigure}[b]{0.45\textwidth}
        \centering
        \includegraphics[width=\textwidth]{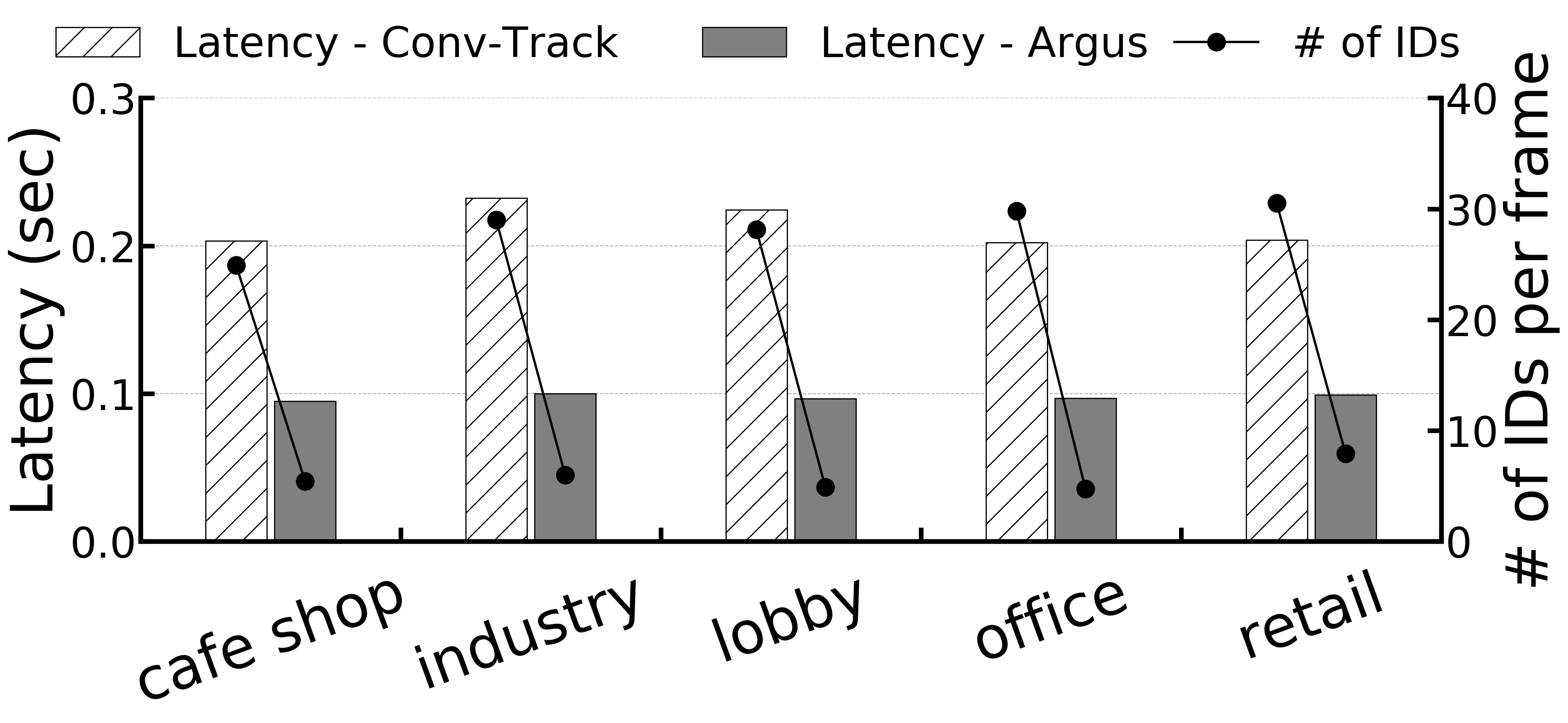}
    	\vspace{-0.2in}
    	\caption{Resource cost.}
    	\label{fig:eval_overall_mmptrack_resource}
    \end{subfigure}
    \hspace{3mm}
    \begin{subfigure}[b]{0.45\textwidth}
		\centering
        \includegraphics[width=\textwidth]{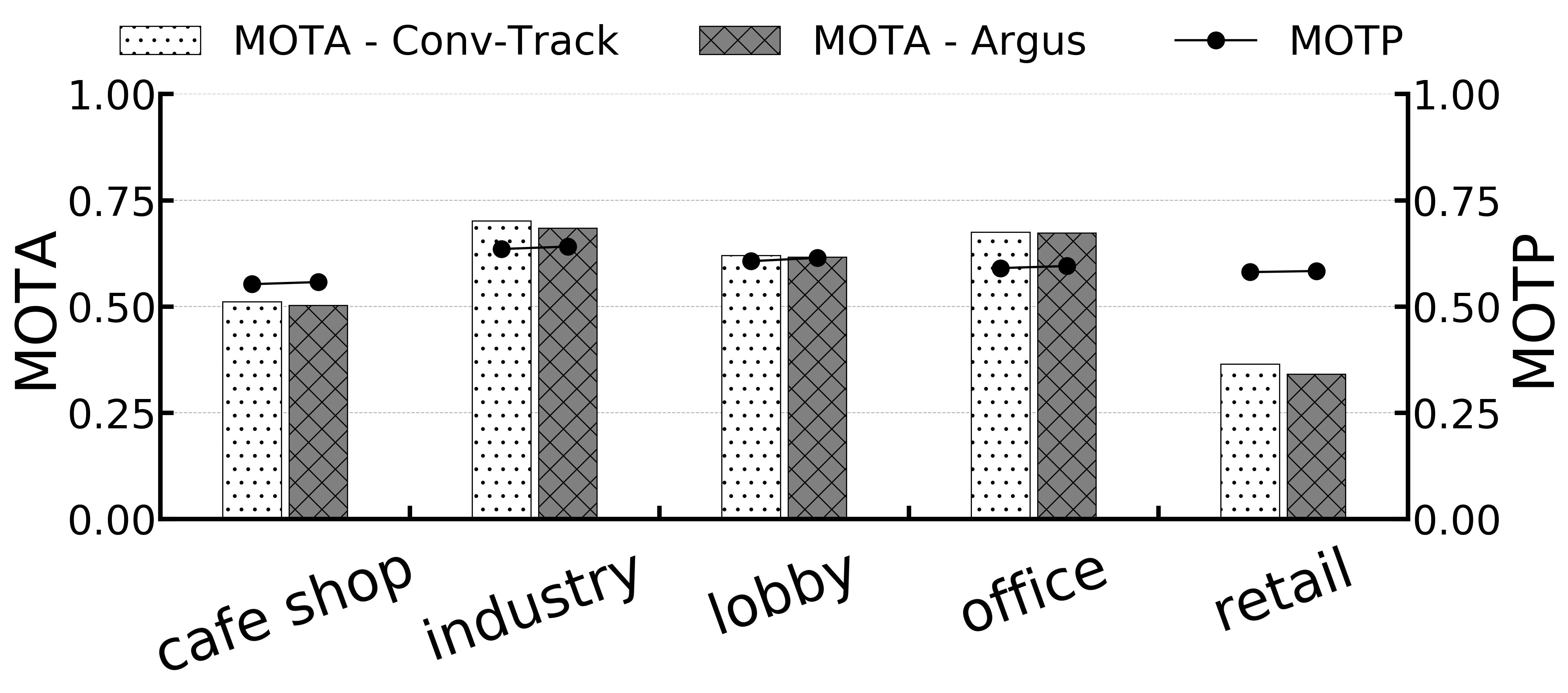}
	    \vspace{-0.2in}
    \caption{Tracking quality.}
    \label{fig:eval_overall_mmptrack_quality}
    \end{subfigure}
    %\vspace{-0.15in}
    \caption{Overall performance on MMPTRACK.}
    \label{fig:eval_overall_mmptrack}
	\endminipage
\vspace{-10pt}
\end{figure*}

\subsection{\revise{Robustness on Large Scale Benchmark}}
\label{subsec:eval-robustness}
\revise{
We perform large-scale evaluation on the MMPTRACK dataset to validate the robustness of \system{}. Figure~\ref{fig:eval_overall_mmptrack} compares the resource cost and tracking quality results of Conv-Track and \system{}; we omit the results of Spatula-Track and CrossRoI-Track as we observe the similar performance trend in the previous experiments in Figures~\ref{fig:eval_overall_aicity} and \ref{fig:eval_overall_campus}.
% we observe similar performance gain of \system{} compared to them (e.g., Spatula-Track achieves marginal resource gain compared to Conv-Track, as most objects appear in all camera throughout the entire video). 
First, Figure~\ref{fig:eval_overall_mmptrack_resource} shows that \system{} achieves 2.19$\times$ latency gain, which mainly comes from reducing the number of IDs from 28.47 (4-8 objects$\times$4-6 cameras) to 5.79. Next, Figure~\ref{fig:eval_overall_mmptrack_quality} shows the tracking quality of Conv-Track and \system{}. The base accuracy of Conv-Track varies across scenes depending on ground truth labeling granularity and detector performance (e.g., \emph{retail} scenes contain a lot of occlusions resulting in detection failure, whereas the ground truth label is provided for all objects regardless of occlusion). \system{} consistently shows marginal accuracy drop compared to Conv-Track, showing the robustness of our spatio-temporal association technique.
}

\subsection{Real-world Case Study \& System Overhead}

\begin{figure}[t]
    \centering
	\minipage{1.0\columnwidth}
    \includegraphics[width=1\textwidth]{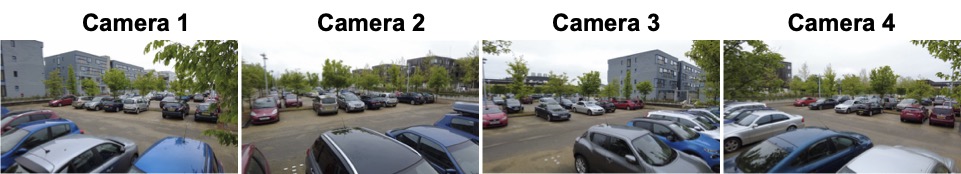}
	\caption{Snapshots of real-world case study.}
    \label{fig:6-eval-case-study-snapshots}
	\endminipage\hfill
\end{figure}

\revise{We conducted a supplementary experiment to investigate both the performance and the operational characteristics of \system{}'s runtime system within a practical deployment scenario. To achieve this objective, we installed four cameras and four Jetson AGX boards within a parking lot of the institute under the consent, employing them to record videos at a resolution of 1080p with a frame rate of 10 frames per second. The parking lot selected for this study spans an approximate area of 50 metres by 25 metres. To ensure a comprehensive coverage, the cameras were positioned at the corners of the parking lot at a height of 3 metres; the corresponding AGX board is connected to the camera via the Ethernet cable and put on the ground. Figure~\ref{fig:6-eval-case-study-snapshots} shows the snapshots of four cameras. Each video stream has an approximate duration of an hour and the ground truth data contains 60 vehicles in total. Since the target objects are vehicles, we used the same object detection model and identification model used in the CityFlowV2 dataset.}

\begin{table}[t]
\centering
%\footnotesize
\small
\caption{Performance of \system{} in the real-world case study.}
\label{tab:eval_case_study}
\begin{adjustbox}{width=0.9\columnwidth}
\begin{tabular}{l|c|c|c|c}
\hline
\multirow{2}{*}{} & \multicolumn{2}{c|}{Resource Efficiency} & \multicolumn{2}{c}{Tracking Quality} \\ \cline{2-5}
                        & Latency      & Number of IDs & MOTP & MOTA            \\ \hline
Parking lot               & 0.21s            & 3.2      & 0.71 & 0.95                         \\ \hline
%AGX              & 0.084s         & 0.038s      & 0.217s & 0.027s                        \\ \hline
\end{tabular}
\end{adjustbox}
%\vspace{-15pt}
\end{table}

\revise{Table~\ref{tab:eval_case_study} shows the \system{}'s overall performance in the real-world case study. It is important to emphasize that we did not conduct a comparative study due to the inability to guarantee consistent behaviours across repetitive experiments in real-world deployment. Moreover, a thorough analysis, compared with baseline methods, has already been reported in preceding subsection. When contrasting the results obtained from the parking lot experiment with the CityFlowV2 dataset, it is interesting to note that the parking lot exhibited marginally superior performance in the aspects both of resource efficiency and tracking quality; we did not compare with the CAMPUS dataset due to the discrepancy in target objects and their respective characteristics. For instance, the average number of identification tasks within the parking lot is 3.2, while the CityFlowV2 dataset showed a higher figure of 10.7. This difference is interesting, especially given that the average number of vehicles captured per video frame in the parking lot exceeded the count of vehicles in the CityFlowV2 dataset. We conjecture this is primarily due to the largely stationary nature of vehicles in the parking lot, allowing the benefit of our spatio-temporal association to be maximized. Similarly, the tracking quality in the parking lot is higher than that in the CityFlowV2 dataset. The MOTP values in the parking lot and the CityFlowV2 dataset were 0.71 and 0.63, respectively. We attribute this to the relatively shorter distance between the camera and the vehicles in the parking lot, enabling the capture of vehicles at a larger scale.}

\revise{We delve deeper into the system overhead of \system{} with this deployment setup. Aside from object detection and identification, the principal operations of the \system{} system encompass two elements: (1) mapping-entry matching and (2) workload distribution decision-making. However, according to our measurements derived from the real-world case study, the overhead associated with both these operations is negligible, quantified as less than a few milliseconds. This minimal overhead can be attributed to our efficient management of mapping entries via a hash table for the first operation. Additionally, for the second operation, the system only needs to consider a relatively small number of cases—typically fewer than five identification operations—when making distribution decisions. This streamlined approach contributes to the overall efficiency and effectiveness of the \system{} system.} 

\subsection{Micro-benchmark}~\label{subsec:overhead}

\begin{table}[t]
\centering
%\footnotesize
\small
\caption{Component-wise microbenchmark.}
\label{tab:eval_benchmark}
\begin{adjustbox}{width=1\columnwidth}
\begin{tabular}{l|c|c|c|c}
\hline
\multirow{2}{*}{Device} & \multicolumn{2}{c|}{Detection (YOLOv5n~\cite{yolov5})} & \multicolumn{2}{c}{Identification (batch size 4)} \\ \cline{2-5}
                        & $1920\times1080$      & $1280\times720$ & ResNet 101~\cite{luo2021empirical} & ResNet 50~\cite{zheng2019joint}            \\ \hline
NX               & 0.359s            & 0.073s      & 0.399s & 0.063s                         \\ \hline
AGX              & 0.084s         & 0.038s      & 0.217s & 0.027s                        \\ \hline
\end{tabular}
\end{adjustbox}
%\vspace{-15pt}
\end{table}

We perform a micro-benchmark to better understand the resource characteristics of model inference on smart cameras. Table~\ref{tab:eval_benchmark} shows the latency of vision models we used on Jetson NX and AGX; we report the detection latency at different image sizes. While the processing capability of smart cameras is still limited compared to the cloud environment, performance can be optimized by applying the right configurations depending on the requirement, e.g., 720p images with people tracking on Jetson NX. We also showed that \system{} can further optimize the latency (and corresponding throughput) by leveraging the spatial and temporal association.

\section{Related Work}~\label{sec:related_work}

\subsection{Cross-Camera Collaboration}

\subsubsection{\revise{Multi-view Tracking using Camera Geometry}}

\revise{Camera geometry, also referred to as the geometry of multiple views, has been studied for multiple decades to enable accurate tracking of objects from different camera views. It deals with the mathematical relationships between 3D world points and their 2D projections onto the image plane~\cite{hartley2003multiple}. By understanding these relationships, the 3D structure of a scene, object, or person has been able to be recovered from multiple 2D views, which enables the tracking of objects even when they move out of one camera's FoV and into another~\cite{szeliski2022computer}. Camera geometry has been applied in various fields, such as robotics, computer vision, and motion capture, where the use of multiple synchronized cameras with overlapping FoVs can improve the tracking accuracy and robustness of the system~\cite{kanatani1993geometric}.}

\revise{The foundation of multi-view tracking is the estimation of the fundamental matrix, which encodes the geometric relationship between the views of two cameras~\cite{hartley2003multiple}. This matrix can be used to compute the epipolar geometry, which describes the relationship between corresponding points in the two images and can be utili`ed to find the corresponding point in the other view when a point is detected in one view~\cite{szeliski2022computer}. By using the fundamental matrix, triangulation techniques can be employed to estimate the 3D position of the tracked object in the scene~\cite{hartley2003multiple}. Also, bundle adjustment, a non-linear optimisation technique, has been used to refine camera parameters and 3D structure of the scene, leading to a more accurate estimation of the object's position~\cite{triggs2000bundle}.}

\revise{Despite the advantages of camera geometry in enabling tracking from multi-camera views, there are several limitations in its deployment. One major challenge is the sensitivity to camera calibration errors, which can lead to inaccurate 3D reconstruction and subsequently impact the tracking performance~\cite{hartley2003multiple}. The calibration process requires the precise estimation of intrinsic camera parameters, such as focal length and lens distortion, and extrinsic parameters, like camera pose and orientation, which can be difficult to obtain in practical applications~\cite{szeliski2022computer}. When using cameras with pan-tilt-zoom (PTZ) capabilities, the camera geometry needs to be recalculated each time the camera view changes, adding to the complexity and computational load of the tracking process. Similarly, the process of calibration should be also repeated each time there are changes in the camera set and topology, such as the addition of a new camera, failure of an existing camera, or the change of a camera's position in a retail store. While \system{} also needs to adapt to these dynamics, it can be done more easily simply by adjusting or regenerating the spatio/temporal association. Moreover, in multi-view tracking using camera geometry, occlusions and ambiguities in object appearances can pose significant challenges in identifying corresponding points across different views, leading to erroneous tracking~\cite{kanatani1993geometric}.} 

\revise{Significantly, calibration may be impossible if video analytics are detached from camera providers. Current video analytics are restricted in leveraging the potential of deployed cameras due to hard-coded analytics capabilities from tightly coupled hardware and software, and isolated camera deployments from various service providers. We propose a paradigm shift towards \textit{software-defined video analytics}, where analytic logics are decoupled from deployed cameras. This allows for dynamic composition and execution of analytic services on demand, without altering or accessing the hardware. For instance, individual shops may wish to run different analytic services using camera streams provided by shopping malls. However, camera parameters may be accessible only to the camera provider (e.g., owner of the shopping mall) and can change without notice depending on the provider's requirements. In contrast, \system{}, relying solely on camera streams, can still be implemented and supported on the video analytics' side.}

\subsubsection{Systems for Cross-Camera Collaboration}

\parjump \noindent \textbf{Enriched video analytics.} One direction for cross-camera collaboration is to provide enriched and combined video analytics from different angles and areas of multiple cameras~\cite{liu2019caesar,li2020cmcaot,jha2021visage}. Liu et al. developed Caesar~\cite{liu2019caesar}, a system that detects cross-camera complex activities, e.g., a person walking in one camera and later talking to another person in another camera, by designing an abstraction for these activities and combining DNN-based activity detection from \emph{non-overlapping} cameras. Li et al. presented a camera collaboration system~\cite{li2020cmcaot} that performs active object tracking by exploiting the intrinsic relationship between camera positions. Jha et al. developed Visage~\cite{jha2021visage} which enables 3D image analytics from multiple video streams from drones. Our work can serve as an underlying on-camera framework for these works, providing multi-camera, multi-task tracking as a primitive task on distributed smart cameras.

\parjump \noindent \textbf{Resource efficiency.} Another direction for cross-camera collaboration is to reduce the computational and communication costs of multiple video streams by exploiting their spatial and temporal correlation~\cite{jain2019scaling,jain2020spatula,guo2021crossroi}. Jain et al. proposed Spatula~\cite{jain2019scaling,jain2020spatula}, a cross-camera collaboration system that targets wide-area camera networks with \emph{non-overlapping} cameras and limits the amount of video data and corresponding communication to be analysed by identifying only those cameras and frames that are likely to contain the target objects. 
\revise{REV~\cite{xu2022rev} also aims at reducing the number of cameras processed by incrementally searching the cameras within the overlapping group and opportunistically skipping processing the rest as soon as the target has been detected.}
\revise{CrossRoI~\cite{guo2021crossroi} and Polly~\cite{li2023polly}} leverages spatial correlation to extract the minimum-sized RoI from overlapping cameras and reduces processing and transmission costs by filtering out unmasked RoIs. 
All such works share the same high-level goal as \system{} in that they leverage spatio-temporal correlation from multiple cameras, but \system{} differs in several aspects, as shown in Table~\ref{tab:comparison}.

\parjump \noindent \textbf{Distributed processing.} There have been several attempts to distribute video analytics workloads from large-scale video data to distributed cameras~\cite{hung2018videoedge,zeng2020distream}. VideoEdge~\cite{hung2018videoedge} optimises the trade-off between resources and accuracy by partitioning the analytics pipeline into hierarchical clusters (cameras, private clusters, public clouds). Distream~\cite{zeng2020distream} adaptively balances workloads across smart cameras. Although this work provided a foundation for the development of distributed video analytics systems, it mainly focused on the video analytics pipeline with one camera as a main workload. In this work, we identify that multi-camera, multi-target tracking is a primary underlying task for overlapping camera environments, and propose an on-camera distributed processing strategy tailed to it.

\subsection{\revise{Resource-Efficient Video Analytics Systems}} 

\noindent \textbf{On-device processing.} Many video analytics systems have been proposed to efficiently process a large volume of video data on low-end cameras, e.g., by adopting on-camera frame filtering~\cite{hsieh2018focus,li2020reducto,xu2021diva}, pipeline adaptation~\cite{apicharttrisorn2019marlin,yi2020eagleeye},
\revise{
edge-cloud collaborative inference~\cite{laskaridis2020spinn, wang2022enabling, almeida2022dyno, kang2017neurosurgeon, yi2020supremo}
RoI extraction~\cite{zhang2021elf, jiang2021remix, yang2022flexpatch, liu2019edge, du2020dds}.
}
On-camera frame filtering techniques filter out the computationally intensive execution of vision models in the early stages, e.g., by dynamically adapting filtering decisions~\cite{li2020reducto} and leveraging cheap CNN classifiers~\cite{hsieh2018focus}. Yi et al. presented EagleEye~\cite{yi2020eagleeye}, a pipeline adaptation system for person identification that selectively uses different face detection models depending on the quality of face images. MARLIN~\cite{apicharttrisorn2019marlin} has been proposed to selectively perform a deep neural network for energy-efficient object tracking. 

\parjump \noindent \textbf{Computation offloading.} 
Several attempts have been made to dynamically adjust video bitrate to optimise the network bandwidth consumption to enable low-latency offloading~\cite{liu2019edge, zhang2018awstream, lu2022gemini}, optimise the video streaming protocol~\cite{du2020dds}, 
\revise{
and design DNN-aware video compression methods~\cite{xie2019grace, du2022accmpeg}. 
}
The other direction for efficient processing is DNN inference scheduling from multiple video streams on the GPU cluster~\cite{zhang2017videostorm,jiang2018chameleon,shen2019nexus}, \revise{DNN merging for memory optimisation~\cite{padmanabhan2023gemel}, privacy-aware video analytics~\cite{cangialosi2022privid, lu2022preva}, and resource-efficient continual learning~\cite{bhardwaj2022ekya, mehrdad2023recl}.
}

While these works manage to achieve remarkable performance improvement, their attempts usually focus on a single camera (or its server). In contrast to these works, we target an environment where multiple cameras are installed in close proximity, and focus on optimising cross-camera operations by leveraging the spatio/temporal association of objects. \system{} can further improve system-wide resource efficiency by applying these techniques.

\section{Discussion and Future Works}
\label{sec:discussion}

\noindent \textbf{Why on-device processing on distributed smart cameras?} 
The cost of video analytics is becoming a huge problem due to the enormous amount of video data. The authors~\cite{xu2021video} studied the six-month deployment of over 1000 cameras at Peking University, China, and reported that the cameras produced over 3 million hours of videos (5.4 PB). If we assume a simple application using the ResNet-18 model at 30 frames per second continuously, the estimated ML operating expenses (OpEx) for six months would be 3.83 million USD if ML inference is executed on the Amazon EC2 server\footnote{100K inferences of ResNet18 costs 0.82 USD and the total number of inferences is 466,560,000,000 (= 30 fps $\times$ 60 sec $\times$ 60 min $\times$ 24 hrs $\times$ 180 days $\times$ 1,000 cameras).}; it will be much higher when the network costs for 5.4 PB of video data are added.
The bigger problem is that, even with excessive cameras, most of the video stream is never used. The study~\cite{xu2021video} further showed that less than 0.005\% of the video data is retrieved and used by less than 2\% of the cameras.

To address this problem, on-camera AI processing is becoming increasingly popular~\cite{zeng2020distream,xu2021case,xu2021video}, enabled by two recent technology trends. First, low-cost, low-power and programmable on-board AI accelerators are becoming available~\cite{antonini2019resource}, such as Nvidia Jetson, Google TPU and Analog MAX78000. Second, lightweight, accurate and robust embedded-ML models are emerging~\cite{gordon2018morphnet, guo2021mistify}. Most importantly, on-device processing is preferred as the privacy-sensitive raw image data does not need to be transferred to the cloud.

\parjump \noindent \textbf{Incorporation with non-overlapping camera collaboration.} 
\system{} currently targets cross-camera collaboration within a closed set of cameras. However, analytics applications might want to track objects in a wide area where large camera networks, including non-overlapping cameras, are installed, e.g., suspect monitoring in a large shopping mall or traffic surveillance in an urban city. Our ultimate goal is to develop a system that supports seamless and efficient tracking across overlapping and non-overlapping cameras by adopting the solutions for non-overlapping cameras~\cite{jain2019scaling,jain2020spatula}.

\parjump \noindent \textbf{Support for diverse coordination topologies.}
For cross-camera coordination, we assume a star topology where the most powerful camera becomes the \emph{head} in a group, scheduling multi-camera multi-target tracking operations for all cameras and the other cameras become group members that follow the head's decision. We believe that our decision is practical because the coordination overhead is marginal as shown in \S\ref{subsec:overhead}, but sophisticated coordination would be necessary if more cameras are involved.

\parjump \noindent \textbf{Further optimization by splitting AI models} \system{} treats AI models as a black box, thereby taking their full execution as a primitive task for distributed processing. Splitting deep neural networks into distributed cameras, e.g., Neurosurgeon~\cite{kang2017neurosurgeon}, would allow further optimisation if we can have access to the weights of the pre-trained models. We leave it as future work.

\parjump \noindent \textbf{Cross-camera communication channel.} For the communication channel between cameras, we consider a Gigabit wired connection, which is already commonly used for existing CCTV networks, e.g., at an intersection~\cite{naphade20215aicity} and a campus~\cite{xu2016campus}. Considering that overlapping cameras are deployed in proximity to each other, such an assumption would be still valid even in other environments. However, when the communication channel is constrained, e.g., over cellular networks, the network overhead may dominate and the latency improvement achieved by multi-camera parallel processing could be less than expected. We leave the detailed analysis as future work. 
\section{Conclusion}
\label{sec:conclusion}

We presented \system{}, a first-kind-of distributed system for robust and low-latency video analytics with cross-camera collaboration on multiple cameras. We developed a novel object-wise spatio-temporal association that optimises the multi-camera, multi-target tracking by intelligently filtering out unnecessary, redundant identification operations. We also developed a distributed scheduling technique that dynamically orders the sequence of camera and bounding box inspection and distributes the identification workload across multiple cameras. Evaluation on three real-world overlapping camera datasets shows that \system{} reduces the number of identification model executions and end-to-end latency by up to 7.13$\times$ and 2.19$\times$ (4.86$\times$ and 1.60$\times$ compared to the state-of-the-arts).

%\begin{acks}
%To Robert, for the bagels and explaining CMYK and color spaces.
%\end{acks}

\bibliographystyle{IEEEtran}
\bibliography{reference.bib}


\end{document}